\pdfoutput=1

\documentclass[11pt]{article}
\usepackage{emnlp2021}
\usepackage{times}
\usepackage{latexsym}

\usepackage[T1]{fontenc}

\usepackage[utf8]{inputenc}

\usepackage{microtype}

\usepackage{graphicx}
\usepackage{subfigure}
\usepackage{multirow}
\usepackage{makecell}
\usepackage{booktabs}

\usepackage{algorithm}
\usepackage{algorithmic}
\usepackage{amsmath}
\usepackage{amsthm}
\usepackage{bm}
\usepackage{ifthen}

\title{{P}arallel Refinements for \\Lexically Constrained 
Text Generation with BART}

\author{Xingwei He \\
  Department of Electrical and Electronic Engineering, \\
  The University of Hong Kong, \\
  Hong Kong, China \\
  \texttt{hexingwei15@gmail.com} \\
  }

\begin{document}
\maketitle
\begin{abstract}
  Lexically constrained text generation aims to control the generated text by 
  incorporating some pre-specified keywords 
  into the output. 
  Previous work injects lexical constraints into the output by controlling the decoding process or refining 
  the candidate output iteratively, 
  which tends to generate generic or ungrammatical sentences, and has high computational complexity. 
  To address these challenges, we propose \textbf{C}onstrained \textbf{BART} (CBART) for lexically constrained text generation.  
  CBART leverages the pre-trained model BART and transfers part of the generation burden from the decoder to the encoder by decomposing this task 
  into two sub-tasks, thereby improving the sentence quality. 
  Concretely, we extend BART by 
  adding a token-level classifier over the encoder, 
  aiming at instructing the decoder where to replace and insert. 
  Guided by the encoder, the decoder refines multiple tokens of 
  the input in one step by inserting tokens before specific positions and re-predicting tokens with low confidence. 
  To further reduce the inference latency, the decoder predicts all tokens in parallel. 
  Experiment results on One-Billion-Word and Yelp show that CBART can generate plausible text with high quality and diversity 
  while significantly accelerating inference. 
\end{abstract}

\section{Introduction} 
Controllable text generation 
aims to generate text in a controlled way, 
such as transferring text style 
\cite{shen2017style,fu2018style,li2018delete,xu2018unpaired} 
and generating text with control codes \cite{Keskar2019CTRLAC}. 
Lexically constrained text generation requires that the given keywords must appear in the output, 
which can be applied to incorporating keywords into a dialog response \cite{mou2016}, 
creating a story with keywords \cite{fan2018hierarchical}, 
generating advertisements for products \cite{miao2019cgmh} and 
writing a concrete meeting summary based on several key phrases. 

To generate sentences with keywords, 
\citet{mou2015backward} proposed a backward and forward language model (B/F-LM).
\citet{Liu2019BFGANBA} applied adversarial learning \cite{goodfellowgenerative} to B/F-LM.  
Both models are limited to generating text with one keyword. 
To incorporate multiple 
keywords into machine translation, 
\citet{Hokamp2017LexicallyCD} proposed grid beam search (GBS) 
by adding an additional constrained dimension to beam search. 
However, GBS 
does not consider the future lexical constraints when generating previous 
tokens, thereby degrading the quality of generated sentences.

\begin{table}[t]
  \scriptsize
    \centering
    \begin{tabular}{
      m{0.05\textwidth}<{\centering}|
      m{0.38\textwidth}<{\raggedright}
      }
      \toprule
      \textbf{Cons} & \textbf{lovely}, \textbf{time}, \textbf{forever}, \textbf{try}\\
      \midrule
      Step 1 & \underline{this} \textbf{lovely} \underline{experience} \textbf{time} \underline{and} \textbf{forever} \underline{to} \textbf{try} \underline{.} \\
      Step 2 & this \underline{was} \textbf{lovely} experience \underline{!} \textbf{time} and \underline{it} \textbf{forever} to \textbf{try} \underline{this} . \\
      Step 3 & this was \underline{a} \textbf{lovely} experience ! \underline{first} \textbf{time} \underline{here} and it \underline{took} \textbf{forever} to \textbf{try} this \underline{place} . \\
      \bottomrule 
    \end{tabular}
    \caption{An example from the Yelp test set demonstrates how CBART generates text with lexical \textbf{cons}traints. 
    In each refinement step, tokens underlined are newly generated. } 
    \label{tab:example}
  \end{table}

Recently, Markov Chain Monte Carlo (MCMC) sampling has been applied to text generation \cite{berglund2015bidirectional,su2018incorporating,Devlin2019BERTPO}. 
Compared with GBS, MCMC-based models can iteratively refine tokens based on contexts. 
CGMH \cite{miao2019cgmh} uses Metropolis-Hastings sampling to generate constrained sentences 
with a series of actions, such as insertion, deletion, and replacement. 
In most cases, refinements conducted by CGMH are invalid because of
the randomly chosen actions and positions. 
To solve this problem, the gradient information \cite{sha-2020-gradient} and 
a token-level classifier \cite{he2021xlentmcmc} are used to determine 
the position to be edited and the action to be taken. 
These models are computation-intensive as they can update only one token in each step. 
POINTER \cite{zhang2020pointer} reduces the inference latency by refining multiple tokens in one step. 
However, POINTER is based on BERT \cite{Devlin2019BERTPO} and imposes all the burden of generation on the decoder. 
Previous work \cite{Lewis2020BARTDS} has shown that 
BART is more suitable than BERT for text generation. 
Nevertheless, BART can not be directly applied to constrained text generation. 
If we feed the keywords into the encoder, the decoder will not guarantee 
that the output contains the given keywords. 

\begin{figure}[!t]
  \centering
    \includegraphics[width=0.45\textwidth]{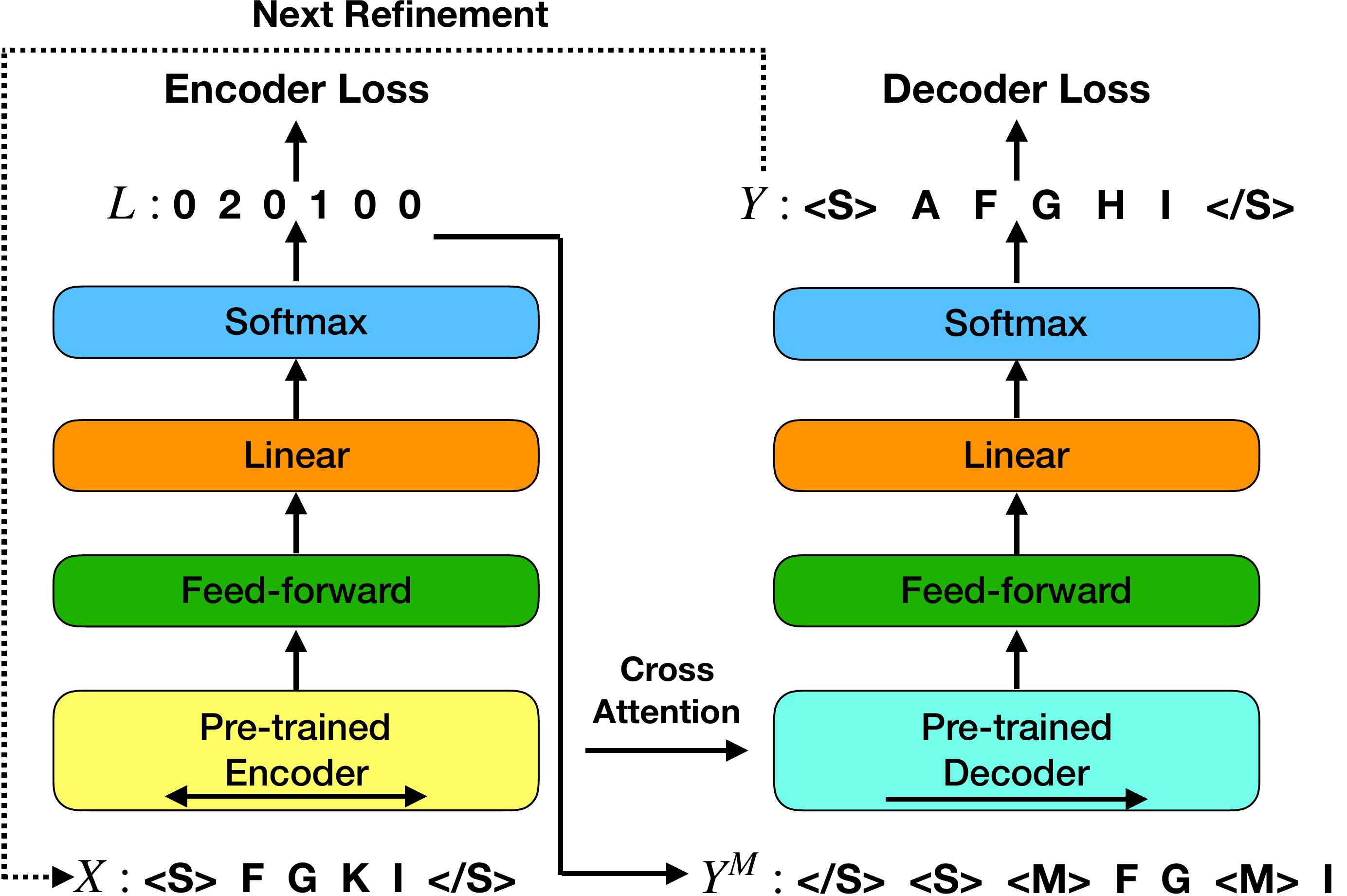} 
  \caption{The overview of our proposed model. $<$S$>$, $<$/S$>$ and $<$M$>$ represent the 
  start, end, and mask tokens. 
  The refinement process indicated by the dashed line only appears in inference. 
  During training, the decoder input $Y^{M}$ is created based on the gold encoder label sequence; 
  while during inference, it is created based on the predicted encoder label sequence.
  }
  \label{framework}
\end{figure}

To alleviate the above problems, we propose CBART, 
a parallel refinement model for lexically constrained text generation, 
shown in Figure \ref{framework}. 
CBART benefits from the large-scale pre-trained model, i.e., BART. 
In addition, CBART shifts part of the decoder's burden to the encoder. 
Specifically, we put a token-level classifier over BART’s encoder, 
in charge of analyzing the input and providing the decoder with coarse-grained modification information, 
such as the positions to be refined and the actions to be conducted. 
The refinement information provided by the encoder enables the decoder to revise multiple tokens of the input 
in one step to make the sentence more fluent, 
such as inserting missing tokens before certain positions and replacing inappropriate tokens with other tokens. 
In addition, the decoder predicts all tokens simultaneously, which further speeds up inference. 
As shown in Table \ref{tab:example}, CBART keeps refining the generated text until completing it 
during inference. 

Our work's main contributions are threefold: 
(1) We propose CBART\footnote{Our code is available at \url{https://github.com/NLPCode/CBART}.} 
for lexically constrained text generation. 
The proposed model takes advantage of the pre-trained model, BART. 
Besides, we lighten the generation burden on the decoder 
by decomposing constrained sentence generation into two sub-tasks. 
Furthermore, CBART can insert and replace multiple tokens in each refinement step and 
predict all tokens in parallel. 
(2) To train CBART, we propose a new method to construct synthetic data. 
(3) Experiment results on One-Billion-Word and Yelp demonstrate
that CBART outperforms previous work in terms of generation quality, 
generation diversity and inference speed. 
\section{Problem Definition}
\textbf{Lexically Constrained Text Generation} aims to incorporate the given keywords 
into the generated text. 
Given a set of lexical constraints, $c_1, c_2, \dots, c_k$, this 
task aims to find a fluent text by maximizing the conditional probability: 
\begin{eqnarray}\label{eq:4}
  X^* = \arg\max_{X} P(X|c_1, c_2, \dots, c_k),
\end{eqnarray} 
where $X$ is the text containing the given keywords. 

\section{Methodology}
In Section \ref{arc}, we will first introduce the proposed model, CBART. 
Then, we will show how to create synthetic datasets to train CBART in Section \ref{syn}. 
Finally, we will introduce several parallel decoding strategies for generating text, 
the repetition penalty mechanism used to discourage the generation of repetitive tokens, 
and the termination criterion for inference in Section \ref{inference}.

\subsection{Model Architecture and Training}\label{arc}
The overview of our proposed model is demonstrated in Figure \ref{framework}. 
The proposed model consists of two modules: 
an encoder $E$ and a decoder $D$. \\
\textbf{Encoder and Action Classifier.} 
We use the pre-trained language model, BART, to initialize the proposed model. 
The encoder is responsible for providing coarse-grained refinement information for the decoder. 
In other words, the encoder is expected to instruct the decoder where to replace and insert. 
To this end, we add a fully connected feed-forward layer over the encoder. More concretely, 
the encoder takes an incomplete sequence as input and outputs a corresponding label sequence. 
Based on the label sequence, the decoder can be aware of how to refine the candidate sentence. 

The encoder serves as a three-class token-level classifier, 
where we use labels $0$, $1$, $2$ to refer to copy, replacement, and insertion actions. 
The copy action means that the decoder should keep the current token.
The replacement action suggests that the decoder should replace the current token
with another token to make the text more coherent. 
Similarly, the insertion action indicates that the decoder 
should insert a token before the current token to complete the text. 
We use $(X,L,Y^{M},Y)$ to represent a training instance. 
$X=\{x_1,\dots,x_n\}$ denotes the incomplete text fed into the encoder 
and $L=\{l_1,\dots,l_n\}$ is the encoder label sequence. 
Then, the cross-entropy loss for the encoder is:
\begin{eqnarray}\label{eq:1}
  L_{encoder} = -\frac{1}{n}\sum_{t=1}^{n}\log_{}p(l_t|x_1,\dots,x_n).
\end{eqnarray} 
\textbf{Decoder.} 
The decoder takes $Y^{M}$ as input and aims to reconstruct the original text $Y$, 
where $Y^{M}=\{y_1^M,\dots,y_m^M\}$ is constructed based on $X$ and $L$,  
and $Y=\{y_1,\dots,y_m\}$ is the decoder label sequence. 
Following BART, the decoder will predict the complete output $Y$ 
rather than the masked tokens of $Y^{M}$ during training. 
We optimize the decoder by minimizing the reconstruction loss:
\begin{eqnarray}\label{eq:2}
  L_{decoder} = -\frac{1}{m}\sum_{t=1}^{m}\log_{}p(y_t|X,y_{\le t}^{M}).
\end{eqnarray} 
\textbf{Joint Training.} 
We jointly optimize the encoder and decoder by minimizing the total loss:
\begin{eqnarray}\label{eq:3}
  L_{total} = L_{encoder}+\alpha L_{decoder}, 
\end{eqnarray} 
where $\alpha$ is the trade-off parameter. 

\subsection{Creating the Synthetic Dataset}\label{syn}
Before training CBART, we need to create the synthetic dataset $D=\{(X,L,Y^{M},Y)\}$.  
Each time, we randomly choose a sentence from One-Billion-Word or Yelp to create the synthetic dataset. 
Suppose the selected text is ``\textit{$<$S$>$ A B C D E F G H I $<$/S$>$}'', 
where $<$S$>$ and $<$/S$>$ denote the start and end tokens. 
Next, we randomly select some tokens from it (e.g., ``\textit{$<$S$>$ F G H I $<$/S$>$}''). 
Then, we randomly replace 15\% of tokens with other tokens (e.g., replace `H' with `K'). 
Therefore, we obtain the encoder input $X=$\{$<$S$>$, F, G, K, I, $<$/S$>$\} and 
the corresponding encoder label sequence $L=$\{0, 2, 0, 1, 0, 0\}, 
where $2$ denotes that a token should be inserted  before `F' and 
$1$ means that `K' should be replaced with another token. 
Further, we construct $Y^{M}=$ \{$<$/S$>$, $<$S$>$, $<$M$>$, F, G, $<$M$>$, I \} 
by inserting a special mask token $<$M$>$ before `F', replacing `K' with $<$M$>$ and 
shifting the sequence one position to the right. 

The decoder label sequence $Y$ is constructed by replacing the masked tokens of $Y^M$ with gold tokens. 
It is trivial to decide the gold token for the replacement action. 
For example, since `H' is replaced with `K', the gold label for the second mask token $<$M$>$ should be `H'. 
However, it is challenging to determine the gold token for the insertion action, 
especially when multiple tokens are missing before a position. 
Since five tokens (`A',`B',`C',`D' and `E') before `F' have been deleted from the original text, 
we need to decide which token should be inserted before `F' first, 
which will be regarded as the gold token for the first mask token $<$M$>$. 

We test five different ways (\textit{Left}, \textit{Middle}, \textit{Right}, \textit{Random} and \textit{TF-IDF})
to construct synthetic datasets for the insertion action. 
The \textit{left} method regards the leftmost token `A' as the first inserted token before `F', so we get 
$Y=$\{ $<$S$>$, A, F, G, H, I, $<$/S$>$\}. 
Similarly, the \textit{middle}, \textit{right}, and \textit{random} methods regard the middle token `C', the rightmost token `E', or 
a randomly chosen token `D' as the first inserted token. 
We also consider the importance of tokens by computing their TF-IDF scores. 
Assume token `B' has the highest TF-IDF value. It will be regarded as the first inserted token.
We conduct an experiment to compare the effect of these methods in Section \ref{analysis}.

\subsection{Inference}\label{inference}
We set lexical constraints as the initial input of the encoder, $X^0$.
We start inference by feeding $X^0$ into the encoder and 
obtain the predicted label sequence $\hat{L}^{0}$ with argmax decoding. 
Next, we construct $Y^{M}$ based on $X^0$ and $\hat{L}^{0}$, and feed it into the decoder. 
Then, we run a decoding strategy on the decoder to get $\hat{Y}^{0}$, using $\hat{Y}^{0}$ as the encoder input of the next refinement step, $X^1$. 
We continue to refine the encoder input, until meeting a termination condition. 
Unlike training, 
we forbid replacing any keyword with the mask token $<$M$>$ when constructing $Y^{M}$, and 
the decoder only needs to predict the masked tokens of $Y^{M}$ to ensure the given keywords appear in the output. 
In the following, we will introduce greedy decoding for the encoder and 
four parallel decoding strategies for the decoder: greedy, 
top-$k$, top-$p$, and multiple-sequence decoding. 
Each decoding strategy allows the decoder to predict all masked tokens of $Y^{M}$ in parallel. \\
\textbf{Greedy Decoding for the Encoder.} 
At the refinement step $r$, the encoder takes $X^r$ as input, 
and we choose the label with the highest probability as the predicted label $\hat{l}_{t}^{r}$
($\hat{L^r}=\{\hat{l}_{1}^{r},\dots,\hat{l}_{n^r}^{r}\}$):
\begin{eqnarray}\label{eq:4}
  \hat{l}_{t}^{r}= \mathop{\arg\max}_{l_t^r}p(l_t^r|X^r). 
\end{eqnarray} \\
\textbf{Greedy Decoding.} Similar to the method mentioned above, greedy decoding selects the token 
with the highest probability for position $t$ as the decoder output ($\hat{Y^r}=\{\hat{y}_{1}^{r},\dots,\hat{y}_{m^{r}-1}^{r},$ $<$/S$>$$\}$): 
\begin{equation}
  \hat{y}_{t}^{r}=
  \begin{cases}
  \mathop{\arg\max}p(y_t^r|X^{r}, y_{\le t}^{M}),& \text{$y_{t+1}^{M}=$ $<$M$>$}\\
  y_{t+1}^{M},& \text{$y_{t+1}^{M}!=$ $<$M$>$}.
  \end{cases}
\end{equation}\label{eq:5} \\
\textbf{Top-$k$ and Top-$p$ Decoding.} 
Since maximization-based decoding, such as greedy decoding and beam search, may cause text degeneration, 
we use top-$k$ \cite{fan2018hierarchical, Holtzman2018LearningTW} 
and top-$p$ decoding 
\cite{Holtzman2020TheCC} to alleviate this problem. 
For each position, top-$k$ decoding samples a token from the $k$ most probable tokens, 
rather than always choosing the most probable one. 
Similarly, top-$p$ decoding samples a token from the smallest possible set of tokens, 
whose cumulative probability exceeds the probability $p$. \\
\textbf{Multiple-sequence Decoding.} 
Top-$k$ or top-$p$ decoding can generate more diverse text 
but risk producing low-quality sentences. 
To remedy this, we propose multiple-sequence decoding. 
When using top-$k$ or top-$p$ decoding to generate sentences, 
we run the decoding method for $N$ times to get multiple sequences. 
Because all sequences are mutually independent, they can be decoded simultaneously. 
After obtaining multiple generated sentences, we resort to the pre-trained language model, 
the GPT-2 small model \cite{Radford2019LanguageMA}, 
to rank these sentences and 
choose the one with the lowest negative log-likelihood (NLL). 
The ranking operation also runs in a non-autoregressive way, 
thus avoiding excessive overhead. \\
\textbf{Repetition Penalty.} 
Even large well-trained generation models might 
generate repetitive phrases or sentences, 
resulting in a lower diversity of the generated text \cite{Holtzman2020TheCC}. 
We find the proposed model suffers from this issue more seriously 
as the masked tokens are predicted conditionally independently in each refinement step. 
To alleviate this problem, we resort to the repetition penalty strategy, 
which discounts the scores of previously generated tokens. 
Slightly different from \citet{Keskar2019CTRLAC}, we discourage the generation of tokens appearing in $Y^{M}$ 
instead of previously generated tokens, achieving the repetition penalty without hurting the non-autoregressive property. 
The probability distribution for the $t$-th token at the refinement step $r$ is defined as follows:
\begin{eqnarray}\label{eq:6}
  p(\hat{y}_{t}^{r}=i) = \frac{exp(h_i/I(i\in Y^M))}{\sum_{j}exp(h_j/I(j\in Y^M))},
\end{eqnarray} 
where $h_i$ is the logit for the $t$-th token. 
If $c$ is true, $I(c)$ equals $\theta$, otherwise equals 1.\\
\textbf{Termination Criterion.} 
During inference, we refine the output token by token. When should we stop refining? 
One method is to monitor the encoder. If all predicted encoder labels are $0$, 
indicating no revision is required by any token, we will stop refining. 
However, this criterion is so strict that refinements may not stop in most cases. 
Therefore, we adopt a relatively loose standard by monitoring the output of the decoder. 
To be specific, if the decoder output is the same as that of the last refinement step, 
we will stop the refinement process. 
\section{Experiments} 
\subsection{Experiment Setup}
\textbf{Datasets and Pre-processing.}
Following \citet{miao2019cgmh} and \citet{zhang2020pointer}, 
we conduct experiments on One-Billion-Word\footnote{http://www.statmt.org/lm-benchmark/} 
and the Yelp dataset\footnote{https://www.yelp.com/dataset}. 
One-Billion-Word is a public dataset for language modeling produced from the \textit{WMT 2011 News Crawl data}.
The Yelp dataset consists of business reviews on Yelp. 
For each dataset, we filter out sentences with length less than $10$ or greater than $40$. 
After preprocessing, we choose $1M$, $0.1M$ sentences from each dataset as the
training and validation sets. We also select $1K$ sentences to provide keywords. 
To be specific, we extract $1$-$6$ keywords from each sentence. Therefore, we 
construct six kinds of test sets for lexically constrained text generation, 
and the size of each test set is $1K$. \\
\textbf{Baselines.}
We compare our proposed model with several strong baselines for lexically constrained text generation, including 
three traditional baselines (sep-B/F, asyn-B/F, and GBS) and 
three recent models (CGMH, POINTER, and X-MCMC-C). 
We implement two variants of the backward and forward language model 
(sep-B/F and asyn-B/F) \cite{mou2015backward}, GBS \cite{Hokamp2017LexicallyCD} and CGMH \cite{miao2019cgmh}. 
For a fair comparison, these baselines are based on the GPT-2 small model
($n_{layer}=12$, $n_{head}=12$, $d_{hidden}=768$, and $117M$ parameters), 
which has a similar architecture to the decoder of BART-large. 

X-MCMC-C benefits from the guidance of the XLNet-based classifier, 
thus substantially improving the generation quality compared to CGMH. 
We train X-MCMC-C with the code provided by \citet{he2021xlentmcmc}, 
which is based on the XLNet-base-cased model ($n_{layer}=12$, $n_{head}=12$, $d_{hidden}=768$, and $110M$ parameters). 
We also compare our model with POINTER \cite{zhang2020pointer}. Similar to our model, POINTER can insert multiple tokens in each step.  
We train two different POINTER models, POINTER and POINTER-2, 
with the code released by \citet{zhang2020pointer}. 
Specifically, POINTER is initialized with BERT-large, while 
POINTER-2 is initialized with the general model, 
pre-trained on the 
 English Wikipedia dataset. 
 Both models have comparable parameters ($n_{layer}=24$, $n_{head}=16$, $d_{hidden}=1024$, and $336M$ parameters) to CBART. 
\\
\textbf{Training and Inference.}
For our model, we create synthetic data with the \textit{left} method. 
For each sentence, we create 10 synthetic data instances. 
Therefore, for each dataset, the size of the synthetic training and validation sets are $10M$ and $1M$. 
We initialize our model with the BART-large model 
($n_{layer}=12$, $n_{head}=16$, $d_{hidden}=1024$, and $406M$ parameters).
We use AdamW \cite{Loshchilov2019DecoupledWD} with an initial learning rate of $1e-5$ and $\alpha=1$ 
to update our proposed model for two epochs 
and choose the checkpoint with the lowest validation loss. 
Please refer to Table \ref{tab:alpha} for the effect of the hyper-parameter $\alpha$. 

During inference, 
we run beam search decoding with beam width = 5 to generate text 
for sep-B/F, asyn-B/F and GBS. Following \citet{he2021xlentmcmc}, we run CGMH and X-MCMC-C for 200 refinement steps  
and select the candidate text with the lowest NLL as output. 
For POINTER, we use greedy decoding to generate constrained text. 
For CBART, we use the four parallel decoding methods (see Section \ref{inference}). 
We apply the repetition penalty to sep-B/F, asyn-B/F, GBS, our models with 
$\theta = 2$, and POINTER with the default value $\theta = 1.25$. 

We implement our model and baselines with HuggingFace \cite{Wolf2019HuggingFacesTS}. 
Results of fine-tuned language models and well-trained classifiers of CBART are shown in the Appendix 
\ref{appendix_a} and \ref{appendix_b}. \\
\textbf{Automatic Evaluation Metrics.} 
We evaluate the generated sentences from two aspects: generation quality and diversity.  
Following previous work \cite{zhang2020pointer}, 
we use BLEU \cite{Papineni2002BleuAM}, NIST \cite{Doddington2002AutomaticEO} 
and METEOR \cite{banerjee2005meteor} as metrics for the generation quality, 
which measure the similarity between the generated text and the human reference. 
A higher BLEU, NIST or METEOR score indicates that a model can generate sentences 
similar to human references. 
In this paper, we do not use NLL as a metric for sentence fluency, 
since a lower NLL value does not always denote better sentence quality. 
Recent work \cite{Holtzman2020TheCC} has found that language models assign low NLL scores not only 
to high-quality sentences, but also to repetitive and generic sentences. 

As for generation diversity, 
we first compute the cumulative 4-gram Self-BLEU score (SB-4) \cite{Zhu2018TexygenAB} 
to measure how similar one sentence is to the other generations 
by treating one sentence as the hypothesis and the others as references. 
Then, we calculate distinct bigrams (D-2) and 4-grams (D-4) \cite{Li2016ADO}, 
which are the number of unique bigrams and 4-grams divided by the total number of generated tokens. 
A lower Self-BLEU or higher distinct n-gram value indicates higher diversity. 
Finally, we measure the n-gram repetitions on a sentence level. 
Since the length of generated sentences varies greatly, we focus on each sentence's first 20 tokens. 
Concretely, if a unigram appears more than two times or a trigram appears more than one time within the first 20 tokens of a sentence, 
we will regard the sentence as containing a repetition.

\begin{table*}[t] 
  \centering
 \scriptsize
   \begin{tabular}{
    m{0.001\textwidth}<{\centering}
    m{0.001\textwidth}<{\centering}
    m{0.1\textwidth}<{\raggedright}|
    m{0.035\textwidth}<{\centering}
    m{0.03\textwidth}<{\centering}
    m{0.035\textwidth}<{\centering}
    m{0.035\textwidth}<{\centering}
    m{0.04\textwidth}<{\centering}|
    m{0.045\textwidth}<{\centering}
    m{0.035\textwidth}<{\centering}
    m{0.04\textwidth}<{\centering}|
    m{0.03\textwidth}<{\centering}
    m{0.035\textwidth}<{\centering}
    m{0.03\textwidth}<{\centering}
    m{0.03\textwidth}<{\centering}
    m{0.03\textwidth}<{\centering}
    }
   \toprule
   &&\multirow{2}{*}{\textbf{Metrics}} & \multicolumn{2}{c} {BLEU $\uparrow$} & \multicolumn{2}{c} {NIST $\uparrow$}& $\uparrow$& $\downarrow$ &\multicolumn{2}{c|} {Distinct $\uparrow$} &$\downarrow$&$\downarrow$ &$\uparrow$& &\\
   &&    &B-2 &B-4 &N-2 &N-4 &M&SB-4 &D-2  &D-4 & Ref & La& S & Rep& Len
    \\
    \midrule
    \multirow{17}{*}{\rotatebox{90}{\textbf{One-Billion-Word}}}&&Human &- &- &-&-&-&10.3\%  &78.1\% &99.5\%  &- &- &- &1.4\% &23.6\\
    \cline{2-16}          
   &\multirow{6}{*}{\rotatebox{90}{\textbf{Baselines}}}&sep-B/F  &4.4\% &0.7\% & 0.616&0.618 &7.0\% &52.1\%   &46.3\% &78.8\% &- &1.900 &5.20 &0.2\% &13.5\\
   &&asyn-B/F  &4.3\% &0.7\%  &0.554 &0.556 &6.8\%  &50.3\% &47.8\% &80.9\% &- &1.865  & 5.29 &0.1\% &13.2 \\
   &&GBS  &10.1\% &2.8\%  & 1.487 &1.497 &13.5\%  &37.0\%    &59.3\% &87.2\%  &- &9.234  &1.07 &0 &14.0\\
   &&CGMH  &9.9\% &3.5\%  &1.153&1.165 &13.1\% &10.2\%   &\textbf{78.9\%} &99.3\%  &200 &9.871  &1.00 &2.5\% &11.6 \\
   &&X-MCMC-C &12.5\% &4.1\% &2.511 &2.532 &13.8\% &16.9\% &69.7\% &98.8\% &200 &31.41&0.31 &1.4\% &16.6\\
   &&POINTER &2.5\% &0.1\%  &0.961&0.961 & 10.2\%&-   &-&-  &6 &- &- &- &65.0 \\
   &&POINTER-2 &8.7\% &1.6\%  & 2.109&2.117 &14.3\% &37.3\%    &46.5\% &90.9\%  &6 &0.727 &13.6 &18.6\% &35.5 \\
   \cline{2-16}                                                 
   &\multirow{5}{*}{\rotatebox{90}{\textbf{CBART}}}&\textbf{Greedy} &15.6\% &\textbf{6.6\%} &2.157&2.191 &\textbf{15.2\%} &22.1\%&66.6\% &97.2\%  &\textbf{4.8} &\textbf{0.351 } &\textbf{28.1} &1.2\% &14.5 \\

   \cline{3-16}
    &&\textbf{$k$=5, $c$=1}   &15.0\% &5.8\%  &2.460&2.491 &14.8\% &16.1\% &70.6\% &98.8\%  &5.1 &0.360  &27.4  &1.0\% & 15.7\\
    &&\textbf{$k$=5, $c$=5}   &15.6\% & 5.9\% &2.677&2.712 &14.9\%  &19.9\%   &65.8\% &98.0\%  &5.3 &0.669  &14.8 &2.1\% &16.4\\
    &&\textbf{$k$=50, $c$=1}    &14.4\%&5.1\%  &2.740&2.768 &14.2\% &11.0\% &76.3\% &99.5\%   &5.5 &0.396  &24.9 &0.8\% &17.2 \\
    &&\textbf{$k$=50, $c$=5}   &15.1\% &5.4\% &2.941&2.974 &14.4\%  &13.7\% &71.6\% &99.2\%  &5.7 &0.720  &13.7 &1.6\% &18.1\\
    
    \cline{3-16}
    &&\textbf{$p$=0.5, $c$=1}   &15.1\% &6.1\% &2.308&2.341 &14.8\% &14.9\% &72.7\% &99.0\%  &5.0 &0.373  &26.5 &0.9\% &15.2\\
    &&\textbf{$p$=0.5, $c$=5}   &\textbf{15.8\%} &6.4\% &2.524&2.561 &14.9\%   &18.2\% &68.0\% &98.4\%  &5.2 &0.674  &14.6 &1.4\% &15.7\\
    &&\textbf{$p$=0.9, $c$=1}   &14.7\% &5.4\% &2.780&2.811 &14.2\%   &\textbf{9.6\%} &78.8\% &\textbf{99.6\%}  &5.5 &0.408  &24.2  &0.7\% &17.3\\
    &&\textbf{$p$=0.9, $c$=5}   &15.2\% &5.5\% &\textbf{2.983}&\textbf{3.017} &14.4\%  &12.2\% &74.4\% &99.4\%  &5.7 &0.759  &13.0 &1.1\% &18.3\\

    \midrule
   \midrule
   \multirow{17}{*}{\rotatebox{90}{\textbf{Yelp}}}&&{Human}  &- &- &- &-&- &26.1\%  &57.7\% &97.0\%  &- &- &- &2.0\% &23.9 \\
   \cline{2-16} 
  &\multirow{6}{*}{\rotatebox{90}{\textbf{Baselines}}}&sep-B/F   &6.9\% &2.1\% &0.521 &0.531 &8.7\% &67.1\%   &31.9\% &64.6\%  &- &1.807  &6.14 &0.1\% &12.6\\
  &&asyn-B/F   &7.5\% &2.3\% &0.698 &0.711  &9.0\% &68.0\% &31.0\%  &64.3\%   &- &1.771  &6.26 &0.1\% &13.4 \\
  &&GBS   &13.6\% &4.5\% &1.680 &1.712  &15.3\%  &59.3\%   &37.5\% &70.2\%  &- &8.634  &1.28 &0.3\% &14.6\\
  &&CGMH   &12.3\% &4.6\% &1.413&1.446 &14.6\%  &23.6\%   &\textbf{60.7\%} &97.7\%  &200 &11.09  &1.00 &5.2\% &12.6\\
  &&X-MCMC-C  & 15.3\% &5.4\% &2.753 &2.803 &15.5\% &38.5\% &47.4\% &92.4\% &200  &31.68 &0.35 &2.0\% &17.8\\
  &&POINTER &4.0\% &0.3\%  & 1.139&1.140 &13.0\% &-    &- &-  &6 &- &- &- &65.3 \\
  &&POINTER-2 &10.6\% &2.4\%  &2.142&2.164 &16.8\% &49.1\%    &35.2\% &86.3\%  &6 &0.741 &15.0 &13.6\% &39.8 \\
  \cline{2-16}                                         
  &\multirow{3}{*}{\rotatebox{90}{\textbf{CBART}}}&\textbf{Greedy}   &19.4\% &\textbf{9.0\%}  &2.541&2.635  &\textbf{17.4\%} &45.1\%&44.4\% &88.1\%  &\textbf{4.9} &0.357  &31.1 &2.3\% &15.4 \\
   \cline{3-16}
   &&\textbf{$k$=5, $c$=1}   &18.4\% &7.4\% &2.946&3.024 &16.8\% &35.8\% &48.7\% &94.1\%  &5.4 &0.391  &28.4 &2.4\% &17.4\\
   &&\textbf{$k$=5, $c$=5}   &19.1\% &7.7\% &3.088&3.172 &16.9\% &38.8\%   &45.9\% &92.7\%  &5.5&0.670  &16.6 &3.5\% &18.0 \\
   &&\textbf{$k$=50, $c$=1}   &17.6\% & 6.7\% &3.103&3.172 &16.2\% &26.4\% &55.9\%   &97.3\%  &5.8 &0.412  &26.9 &1.2\% &19.5 \\
   &&\textbf{$k$=50, $c$=5}   &18.3\% &7.0\%  &3.220&3.292 &16.4\% &29.5\% &52.6\% &96.4\% &6.0 &0.742  &14.9 &2.0\% &20.4  \\
   \cline{3-16}
   &&\textbf{$p$=0.5, $c$=1}   &19.0\% &8.3\% &2.779&2.867  &17.0\% &37.9\% &48.5\% &92.9\% &5.1 &\textbf{0.349}  &\textbf{31.8} &2.0\% &16.5\\
   &&\textbf{$p$=0.5, $c$=5}   &\textbf{19.7\%} &8.6\% &2.985&3.082  &17.1\% &41.2\%   &45.4\% &91.2\% &5.3 &0.658 &16.9 &3.3\% &17.2 \\
   &&\textbf{$p$=0.95, $c$=1}   &17.6\% &6.8\% &3.163&3.234 &16.0\% &\textbf{23.5\%}   &59.3\% &\textbf{97.9\%}  &5.9 &0.384  &28.9 &1.3\% &20.2 \\
   &&\textbf{$p$=0.95, $c$=5}   &18.2\% &6.9\% &\textbf{3.225}&\textbf{3.295} &16.3\% &27.3\%   &55.2\% &97.0\%  &6.1 &0.703 &15.8 &2.1\% &21.1\\
   \bottomrule
 \end{tabular}
 \caption{Results on One-Billion-Word and Yelp test sets. 
 (``Human'' means human references. 
 $k$ and $p$ are hyper-parameters for top-$k$ and top-$p$ decoding, respectively.
 $c$ is the number of parallel sequences for the multiple-sequence decoding. 
 ``M'' refers to METEOR. 
 ``Ref'' denotes the average number of refinements taken during decoding. 
 ``La'' (latency) is the average decoding time (second) per sentence computed on test sets without mini-batching. 
 ``S'' denotes speedup. ``Rep'' means the percentage of sentences containing n-gram repetitions. 
 ``Len'' represents the average length of the generated sentences.)
 Results for sep-B/F and asyn-B/F are on the test set with $N=1$ constraint. 
 Results for the remaining models are averaged over the six test sets with $N=1$ to $N=6$ lexical constraints. 
 }\label{tab:result2} 
\end{table*}

\subsection{Main Comparison Experiment Results} 
We show the experiment results on One-Billion-Word and Yelp test sets in Table \ref{tab:result2}, 
from which we can draw four conclusions: \\
(1) \textbf{Sep-B/F, asyn-B/F and GBS have low generation quality and diversity. }
Sep-B/F, asyn-B/F and GBS have low BLEU, NIST and METEOR values, 
indicating poor generation quality. 
That is possible because these models force keywords to be incorporated 
into outputs during decoding, thus degrading the generation quality. 
Moreover, compared with human-written text, sentences generated by sep-B/F, asyn-B/F and GBS 
are much less diverse as they have higher Self-BLEU and lower distinct n-gram scores. 
Since these models are not aware of keywords before generation, 
they tend to generate generic phrases (\textit{`he said'}, \textit{`he would'}, etc.), 
thereby degrading the generation diversity. \\
(2) \textbf{Sampling-based methods have high generation diversity and low quality. }
CGMH is on par with humans in generation diversity (Self-BLEU and distinct n-gram scores), 
yet it comes at the expense of degrading sentence quality (lower BLEU scores). 
This conclusion is in line with the results 
in previous work \cite{zhang2020pointer, he2021xlentmcmc}. 
Compared with CGMH, X-MCMC-C slightly boosts the generation quality 
due to the decreasing of random modifications. \\
(3) \textbf{POINTER reduces the inference latency yet with low generation quality. } 
Compared with other baselines, 
POINTER significantly reduces the inference latency but has low generation quality. 
There are two possible reasons: 
(1) POINTER is based on BERT, which is not designed for text generation; 
(2) POINTER imposes all the burden of generation on the decoder. 
In addition, pre-training POINTER on Wikipedia (POINTER-2) improves the performance, 
consistent with what is observed in previous work \cite{zhang2020pointer}. 
However, it is not fair for other models, 
as we may also make improvements by training them on larger datasets. \\
(4) \textbf{The proposed model outperforms baselines in most metrics. }
Similar to POINTER, CBART can refine multiple tokens in each refinement step. 
Therefore, CBART only needs several steps to complete a sentence with the given keywords, 
thus dramatically reducing the inference time. 
As shown in Table \ref{tab:result2}, CBART with greedy decoding needs around 
five refinement steps and is about 28 and 31 times faster 
than CGMH on One-Billion-Word and Yelp. 
However, CBART outperforms POINTER in generation quality and diversity by a large margin. 
This is possible because CBART shifts part of the burden from the decoder to the encoder and
 benefits from the intrinsic generation ability of BART. 

To summarize, it is non-trivial to satisfy all metrics when generating constrained sentences. 
CBART with greedy decoding can generate sentences with relatively high sentence quality and diversity  
while largely reducing the inference latency. 
We can also control sentence quality and diversity with top-$k$ or top-$p$ sampling. 
For example, increasing $k$ or $p$ allows CBART to generate tokens with low probabilities, 
thus improving sentence diversity. 

\begin{table}[t] 
  \scriptsize
    \centering
      \begin{tabular}{
       m{0.04\textwidth}<{\centering}
       m{0.06\textwidth}<{\centering}|
       m{0.1\textwidth}<{\centering}|
       m{0.04\textwidth}<{\centering}
       m{0.08\textwidth}<{\centering}
       }
  
      \toprule
      \multicolumn{5}{c}{\textbf{Fluency}: which sentence is more fluent?}\\
      \midrule
      \multicolumn{2}{c|}{Model A won}& Tied& \multicolumn{2}{c}{Model B won}\\
      \midrule
      CBART&32.7\% & 24.0\% &\textbf{43.3\%}&Human\\
      \midrule
      CBART&\textbf{70.0\%} & 14.0\%& 16.0\% &GBS\\
      \midrule
      CBART&\textbf{56.0\%} &25.3\% &18.7\%   &CGMH\\
      \midrule
      CBART& \textbf{44.7\%}&33.3\% & 22.0\% &X-MCMC-C\\
      \midrule
      CBART& \textbf{77.3\%}& 18.0\% & 4.7\% &POINTER-2\\
      
      \midrule
      \multicolumn{5}{c}{\textbf{Informativeness}: which sentence is more informative?}\\
      \midrule
      \multicolumn{2}{c|}{Model A won}& Tied& \multicolumn{2}{c}{Model B won}\\
      \midrule
      CBART& 8.7\%& 9.3\% &\textbf{82.0\%} &Human\\
      \midrule
      CBART&\textbf{65.3\%} & 10.7\% & 24.0\%&GBS\\
      \midrule
      CBART& \textbf{52.7\%}&18.0\% & 29.3\%   &CGMH\\
      \midrule

      CBART& \textbf{33.3\%}&35.4\% & 31.3\% &X-MCMC-C\\
      \midrule
      CBART& \textbf{48.0\%}& 12.0\% &40.0\% &POINTER-2\\
  
      \bottomrule
    \end{tabular}
    \caption{Human evaluation on One-Billion-Word. 
    }\label{tab:result3}
  \end{table}
\subsection{Human Evaluation} 
To further assess the proposed model, we conduct a human evaluation. 
We compare CBART (greedy decoding) with GBS, CGMH, X-MCMC-C, POINTER-2 and human references. 
For each model, we randomly select 50 sentences and 
invite three volunteers\footnote{All annotators have Bachelor's or higher degrees. They are independent of our research group.}
 to compare the sentences generated by different models. 
Following previous work \cite{huang-etal-2020-inset}, each annotator should compare sentence A with sentence B 
and decide which one is more fluent or informative. 
A tie is allowed if they have no preference. 
Sentences in each pair are shuffled before annotation to avoid bias. 
We show the results of the human evaluation in Table \ref{tab:result3}. 
The inter-rater agreement measured by Fleiss' kappa \cite{Fleiss1971MeasuringNS} 
is 0.69 and 0.65 for fluency and informativeness, 
indicating a substantial inter-rater agreement, according to \citet{Landis1977TheMO}.
CBART outperforms baselines in both fluency and informativeness. 
CBART is even on par with humans in terms of sentence fluency. 
However, the proposed model still lags far behind humans in terms of informativeness. 
We speculate that this is because the model is only taught 
to generate fluent sentences during training, 
resulting in the generated sentences being shorter than human references. 
\begin{table}[t] 
  \scriptsize
    \centering
      \begin{tabular}{
       m{0.01\textwidth}<{\centering}|
       m{0.07\textwidth}<{\raggedright}|
       m{0.035\textwidth}<{\centering}
       m{0.035\textwidth}<{\centering}
       m{0.035\textwidth}<{\centering}
       m{0.035\textwidth}<{\centering}
       m{0.041\textwidth}<{\centering}
       m{0.035\textwidth}<{\centering}
       }
  
      \toprule
      \multirow{2}{*}{\#}&\textbf{Metrics}  & $\uparrow$& $\uparrow$ & $\uparrow$ & $\downarrow$ &\\
      &\textbf{CBART} &B-2 &N-2 &M &SB-4&Rep\\
      \midrule
      1&Left&\textbf{19.4\%}&\textbf{2.541}&\textbf{17.4\%}&45.1\%&2.3\%\\
      2&Middle&19.3\%&2.431&\textbf{17.4\%}&45.5\%&3.1\%\\
      3&Right&18.5\%&2.151&17.2\%&43.9\%&2.5\%\\
      4&Random&18.2\%&2.030&17.1\%&42.2\%& 2.1\%\\
      5&TF-IDF&16.7\%& 1.778&16.7\%&\textbf{36.7\%}&0.9\%\\
      \midrule
      6&w/o RP&\textbf{22.3\%}&\textbf{3.537}&17.2\%&50.7\% &54.7\%\\
      \midrule
      7&$N$ = 1&  5.7\%&0.312&8.3\% &55.2\% &0.3\%\\
      8&$N$ = 2&  9.7\%&0.742&11.8\% &52.3\% &0.7\%\\
      9&$N$ = 3&  16.0\%&1.726&15.7\% &45.5\% &1.8\%\\
      10&$N$ = 4&  22.4\%&2.996&19.4\%&42.2\% &2.8\%\\
      11&$N$ = 5&  28.0\%&4.122&22.7\%&38.7\% &3.6\%\\
      12&$N$ = 6&  \textbf{34.5\%}&\textbf{5.347}&\textbf{26.3\%}& \textbf{36.5\%}&4.7\%\\
    \bottomrule
    \end{tabular}
    \caption{ Results of CBART variants on Yelp test sets. 
      At the top, we compare the performance of CBART trained with different synthetic datasets. 
    At the middle, we show the impact of the repetition penalty (RP).
    At the bottom, we show the effect of the number of lexical constraints $N$. 
    }\label{tab:result4}
\end{table}
\subsection{Ablation Study and Analysis}\label{analysis}
\textbf{Effect of Methods for Creating Synthetic Datasets.} 
We train CBART on different synthetic datasets constructed with different insertion ways 
and show the results at the top of Table \ref{tab:result4}. 
CBART trained with the \textit{left} method gets the best sentence quality in terms of BLEU, NIST and METEOR. 
We speculate that CBART fine-tuned with the \textit{left} method generates the leftmost tokens first, 
which is consistent with the generation order of BART. 
Therefore, CBART trained with the \textit{left} method has a relatively smaller gap between training and fine-tuning  
than other variants. \\
\textbf{Effect of the Repetition Penalty.} 
In Table \ref{tab:result4}, CBART with the repetition penalty removed tends to get stuck in repetition loops, 
and the percentage of sentences containing repetitions surges from $2.3\%$ to $54.7\%$
(row 1 vs. row 6). \\
\textbf{Effect of the Number of Constraints.}
As shown at the bottom of Table \ref{tab:result4}, with constraints increased, 
BLEU, NIST, and METEOR also increase. 
Since the increased constraints shrink the solution space, 
it becomes much easier for the model to generate sentences close to human 
references, thus boosting the generation quality. 
Furthermore, the model diversity improves 
because it is less likely to generate similar sentences as constraints increase. 
\begin{table}[t] 
    \scriptsize
      \centering
        \begin{tabular}{
         m{0.01\textwidth}<{\centering}|
         m{0.12\textwidth}<{\raggedright}|
         m{0.033\textwidth}<{\centering}
         m{0.033\textwidth}<{\centering}
         m{0.033\textwidth}<{\centering}
         m{0.033\textwidth}<{\centering}
         m{0.033\textwidth}<{\centering}
         }
    
        \toprule
        \multirow{2}{*}{\#}&\textbf{Metrics} & $\uparrow$& $\uparrow$ & $\uparrow$ &$\downarrow$& \\
        &\textbf{CBART}  &B-2 &N-2 &M &SB-4&Rep\\
        \midrule
        1&LM w/ CAM& \textbf{19.4}\%&\textbf{2.541}&\textbf{17.4\%}& 45.1\% &2.3\%\\
        2&MLM w/ CAM& 18.7\%&2.296&17.2\% &\textbf{43.9\%}  &2.1\%\\
        3&LM w/o CAM& 19.2\%&2.453&17.4\% &45.8\%   &2.5\%\\
        4&MLM w/o CAM& 10.2\%&1.835&13.1\% &53.3\%  &99.3\%\\
        \midrule
         5& w/ Random& 16.4\%&1.769&16.5\% &47.9\%  &3.6\%\\
         6 &w/ BART-base& 18.4\%&2.310&17.0\%& 45.6\% & 1.9\%\\
         7& w/ BART-large&\textbf{19.4}\%&\textbf{2.541}&\textbf{17.4\%}& \textbf{45.1\%} &2.3\%\\
      \bottomrule
      \end{tabular}
      \caption{ Results of CBART variants on Yelp test sets.  At the top, we show the effect of training objectives. 
      At the bottom, we compare the effect of pre-trained models.  
      (``CAM'' is the causal attention mask. 
      ) 
      }\label{tab:result5}
\end{table}
\\
\textbf{Effect of Training Objectives.} 
We conduct experiments to analyze the effect of training CBART with different training objectives, 
language modeling (LM) and masked language modeling (MLM). 
The difference between them is that 
during training, LM will reconstruct the original text, 
while MLM only predicts the masked tokens. 
As shown in Table \ref{tab:result5}, 
CBART trained with LM (row 1) performs better than CBART trained with MLM (row 2), 
possibly because LM is more suitable for text generation, in line with previous results \cite{Lewis2020BARTDS}. 

Previous non-autoregressive translation (NAT) models \cite{Ghazvininejad2019MaskPredictPD,Stern2019InsertionTF} 
removed the causal attention mask (CAM) from the decoder so that each target token can attend to other tokens of the decoder input. 
Therefore, we also conduct experiments to analyze the effect of CAM. 
However, we do not observe any significant improvement (row 1 vs. row 3) for our task. 
This arises from the difference in input between our task and machine translation. 
For machine translation, the encoder and decoder take different languages as input. 
By comparison, for our task, 
the decoder input is constructed by 
inserting mask tokens before some positions and replacing some tokens with mask tokens. 
Therefore, 
each token of the decoder input $Y^M$ also appears in the encoder input $X$. 
When training CBART with CAM, each token of the decoder input can attend to the encoder input via cross attention, 
which is equivalent to attending to other tokens of the decoder input by removing CAM. 
That is why we cannot make further improvements by removing CAM. \\
\textbf{Effect of Pre-trained Models.}
We train two base CBART models with 6 layers initialized with 
random values (row 5) or the BART-base model (row 6) and a large CBART model initialized 
with the BART-large model (row 7). 
From Table \ref{tab:result5}, we conclude that pre-trained models (row 5 vs. row 6)
and model size are important (row 6 vs. row 7). 
These are in line with our intuitions: (1) CBART inherits some syntactic and semantic knowledge from BART;
(2) increasing the model size will improve the performance. 
Note that in this paper, the experiment results of our proposed model are based on the large CBART model, if not specified. 
\textbf{Effect of the Hyper-parameter $\alpha$.}
Since $\alpha$ is an important hyper-parameter for training CBART, 
we train CBART-base (the base CBART model, which is initialized with BART-base.) with different $\alpha$. 
From the results in Table \ref{tab:alpha}, we find that $\alpha=1.0$ is an appropriate value for CBART. 
We speculate that when $\alpha$ is too large, CBART will pay more attention to the generation/decoder loss. 
On the contrary, when $\alpha$ is too small, 
CBART will put more focus on the classification/encoder loss.
Both losses are essential for the performance of CBART. 
As we can see, CBART trained with $\alpha=1.0$ performs best in most metrics, 
which is a good trade-off between the encoder loss and the decoder loss. 
\begin{table}[t] 
\footnotesize
  \centering
    \begin{tabular}{
     m{0.01\textwidth}<{\raggedright}|
     m{0.07\textwidth}<{\raggedright}|
     m{0.035\textwidth}<{\centering}
     m{0.035\textwidth}<{\centering}
     m{0.035\textwidth}<{\centering}
     m{0.045\textwidth}<{\centering}
     m{0.035\textwidth}<{\centering}
     }

    \toprule
    \multirow{2}{*}{\#}&\multirow{2}{*}{Metrics} &  $\uparrow$& $\uparrow$ & $\uparrow$ & $\downarrow$ &\\
    & &B-2 &N-2 &M &SB-4&Rep\\
    \midrule
    1&$\alpha=0.5$&  17.7\%&2.024& 16.9\% &\textbf{44.4\%}& 2.1\%\\
    2&$\alpha=1.0$& \textbf{18.4\%}&\textbf{2.310}& \textbf{17.0\%}&45.6\%& 1.9\%\\
    3&$\alpha=1.5$&18.2\%&2.203& 17.0\% &45.3\%& 2.4\% \\
    4&$\alpha=2.0$&18.1\%&2.144& 17.0\% &44.8\%& 2.0\%\\
    5&$\alpha=2.5$&18.1\%& 2.139& 17.0\% &45.2\%& 2.1\%\\
    6&$\alpha=3.0$&18.1\%& 2.155& 17.0\% &44.8\%& 2.0\%\\
    \bottomrule
  \end{tabular}
  \caption{The effect of $\alpha$. Results of CBART-base trained with different $\alpha$ on the Yelp test set. 
  }\label{tab:alpha}
\end{table}

\subsection{Samples and Analysis}
We show some sentences generated by baselines and our proposed model 
with lexical constraints extracted from Yelp test sets in Table \ref{tab:case1}. 
From this table, we can see that 
the sentence generated by CBART with greedy decoding is more fluent and meaningful than baselines. 
We can also generate more diverse and informative sentences with top-$p$ or top-$k$ sampling 
by increasing $k$ or $p$, 
but there is a risk of generating less fluent sentences (see $p=0.95,c=1$). 
Increasing the number of sampling sequence $c$ can slightly alleviate this (see $p=0.95,c=5$). 
More generated sentences are shown 
in Table \ref{tab:case2} 
and Table \ref{tab:case3} in the Appendix \ref{appendix_d}.

\begin{table}[t]
  \scriptsize
    \centering
    \begin{tabular}
      {
      m{0.042\textwidth}<{\raggedright}|
      m{0.388\textwidth}<{}
      }
      \toprule
      \textbf{Cons} & \textbf{family}, \textbf{good}, \textbf{location}, \textbf{star}\\
      \midrule
      \textbf{Human}& my \textbf{family} and i did not have a \textbf{good} experience here at this \textbf{location} . the one \textbf{star} is for the food .\\
      \midrule
      \textbf{GBS} & this is a \textbf{good} \textbf{location} for \textbf{family} and \textbf{star} wars fans .\\
      \midrule
      \textbf{CGMH}& very nice \textbf{family} friendly spot , \textbf{good} \textbf{location} on \textbf{star} .\\
      \midrule
      \multicolumn{2}{l}{\textbf{CBART with Different Decoding Methods}}\\
      \midrule
      \textbf{greedy}& my \textbf{family} and i always get \textbf{good} food at this \textbf{location} . the one \textbf{star} is for customer service .\\
      \midrule
      \textbf{$k$=5, $c$=1}&my favorite \textbf{family} owned restaurant ! friendly staff , \textbf{good}  food and the \textbf{location} is great and always a 5 \textbf{star} experience .	\\
      \midrule
      \textbf{$k$=50, $c$=1}&it was a \textbf{family} friendly restaurant in las vegas with \textbf{good}  food for children and nice \textbf{location} , this two \textbf{star} is because of service .	\\
      \midrule
      \textbf{$k$=50, $c$=5}&great \textbf{family} owned restaurant and \textbf{good} food . great \textbf{location} and a solid \textbf{star} coffee experience in general !	\\
      \midrule
      \textbf{$p$=0.5, $c$=1}&my \textbf{family} and i really enjoyed this place . \textbf{good} food and great \textbf{location} , but my one \textbf{star} rating is for the service !	\\
      \midrule
      \textbf{$p$=0.95, $c$=1}&olive case has been \textbf{family} run and had a \textbf{good} reputation , but this \textbf{location} was the \textbf{star} . if you purchase online ask to check in on the map !	\\
      \midrule
      \textbf{$p$=0.95, $c$=5}&our \textbf{family} went here for dinner . we always have such an \textbf{good} time at this \textbf{location} on a \textbf{star} wars movie night !	\\
      \bottomrule
    \end{tabular}
    \caption{Sentences generated by baselines and CBART with lexical \textbf{cons}traints extracted from Yelp test sets. ``Human'' refers to the human reference.}\label{tab:case1}
  \end{table}

\section{Related Work}
\textbf{Pre-trained Language Model.} 
Large-scale pre-trained models have achieved remarkable success in 
many natural language understanding (NLU) tasks \cite{Devlin2019BERTPO,Liu2019RoBERTaAR} and 
natural language generation (NLG) tasks \cite{Radford2018ImprovingLU,Radford2019LanguageMA}. 
Some unified pre-trained language models, such as XLNet \cite{Yang2019XLNetGA}, 
UNILM \cite{Dong2019UnifiedLM} and BART \cite{Lewis2020BARTDS}, have attempted to solve both NLU and NLG tasks. 
Unlike previous work, we have extended our work from BART, in order to generate text under specified lexical constraints. 
\\
\textbf{Non-autoregressive Generation.} 
NAT \cite{Gu2018NonAutoregressiveNM} generates all tokens in parallel, thus speeding up the inference. 
The parallel decoding comes at the cost of degrading generation quality since NAT breaks the dependency among target tokens. 
To alleviate this problem, recent NAT models \cite{Lee2018DeterministicNN,Stern2019InsertionTF, gu2019levenshtein, Ghazvininejad2019MaskPredictPD} generate the output with several steps, making a trade-off 
between the decoding speed and generation quality. 
\citet{susanto-etal-2020-lexically} extended Levenshtein Transformer \cite{gu2019levenshtein} by
injecting keywords into machine translation.
It works on machine translation because the source and target are mostly aligned 
and the solution space is small. 
However, it performs worse than BART on general text generation \cite{lin-etal-2020-commongen},  
which has a much larger solution space. 
Different from previous work, CBART inherits the generation ability from BART while maintaining the decoding speed of NAT models. \\
\textbf{Lexically Constrained Text Generation.} 
B/F-LMs \cite{mou2015backward,Liu2019BFGANBA} are limited to generating text with one lexical constraint. 
GBS \cite{Hokamp2017LexicallyCD,post-vilar-2018-fast} incorporates multiple constraints into the output 
by controlling the decoding process yet degrades the generation quality and diversity. 
Recently, MCMC sampling has been applied to constrained text generation \cite{miao2019cgmh, zhang2020languagegeneration}. 
Nevertheless, refinements conducted by these models are decided randomly. 
To alleviate this, \citet{sha-2020-gradient} used the gradient information, and \citet{he2021xlentmcmc} used a token-level classifier 
to decide the refinements, 
but these models updated only one token in each step. 
POINTER \cite{zhang2020pointer} can refine multiple tokens in each step 
but is based on BERT and imposes all the generation burden on the decoder, thus degrading the sentence quality. 
To solve these problems, we propose CBART, which is based on BART and transfers part of the decoder's burden to the encoder. 

Another approach also generates text based on keywords \cite{fan2018hierarchical, tan2020progressive, lin-etal-2020-commongen} 
but does not force them to appear in the output. 
By comparison, our work focuses on requiring all keywords to appear in the output. 
\section{Conclusion}
In this paper, we presented CBART 
for lexically constrained text generation.
Compared with previous work, CBART leverages BART and transfers part of the burden from the decoder to the encoder. 
Furthermore, CBART refines multiple tokens in parallel in each refinement step, 
thus accelerating the inference. 
Experiment results on One-Billion-Word and Yelp datasets show that CBART can generate fluent and diverse text with lexical constraints
 and dramatically reduce the inference time. 
\section*{Acknowledgments}
We would like to thank the anonymous reviewers for their constructive and informative feedback. 

\bibliographystyle{acl_natbib}
\bibliography{emnlp2021}

\begin{thebibliography}{46}
\expandafter\ifx\csname natexlab\endcsname\relax\def\natexlab#1{#1}\fi

\bibitem[{Banerjee and Lavie(2005)}]{banerjee2005meteor}
Satanjeev Banerjee and Alon Lavie. 2005.
\newblock \href {https://aclanthology.org/W05-0909} {{METEOR}: An automatic
  metric for {MT} evaluation with improved correlation with human judgments}.
\newblock In \emph{Proceedings of the {ACL} Workshop on Intrinsic and Extrinsic
  Evaluation Measures for Machine Translation and/or Summarization}, pages
  65--72.

\bibitem[{Berglund et~al.(2015)Berglund, Raiko, Honkala, K{\"a}rkk{\"a}inen,
  Vetek, and Karhunen}]{berglund2015bidirectional}
Mathias Berglund, Tapani Raiko, Mikko Honkala, Leo K{\"a}rkk{\"a}inen, Akos
  Vetek, and Juha~T Karhunen. 2015.
\newblock \href
  {https://proceedings.neurips.cc/paper/2015/hash/c75b6f114c23a4d7ea11331e7c00e73c-Abstract.html}
  {Bidirectional recurrent neural networks as generative models}.
\newblock In \emph{NIPS}, pages 856--864.

\bibitem[{Devlin et~al.(2019)Devlin, Chang, Lee, and
  Toutanova}]{Devlin2019BERTPO}
Jacob Devlin, Ming-Wei Chang, Kenton Lee, and Kristina Toutanova. 2019.
\newblock \href {https://aclanthology.org/N19-1423} {{BERT}: Pre-training of
  deep bidirectional transformers for language understanding}.
\newblock In \emph{Proceedings of NAACL-HLT}, pages 4171--4186.

\bibitem[{Doddington(2002)}]{Doddington2002AutomaticEO}
George Doddington. 2002.
\newblock \href {https://dl.acm.org/doi/abs/10.5555/1289189.1289273} {Automatic
  evaluation of machine translation quality using n-gram co-occurrence
  statistics}.
\newblock In \emph{Proceedings of the second international conference on Human
  Language Technology Research}, pages 138--145.

\bibitem[{Dong et~al.(2019)Dong, Yang, Wang, Wei, Liu, Wang, Gao, Zhou, and
  Hon}]{Dong2019UnifiedLM}
Li~Dong, Nan Yang, Wenhui Wang, Furu Wei, Xiaodong Liu, Yu~Wang, Jianfeng Gao,
  Ming Zhou, and Hsiao-Wuen Hon. 2019.
\newblock \href
  {https://proceedings.neurips.cc/paper/2019/hash/c20bb2d9a50d5ac1f713f8b34d9aac5a-Abstract.html}
  {Unified language model pre-training for natural language understanding and
  generation}.
\newblock In \emph{NIPS}, pages 13063--13075.

\bibitem[{Fan et~al.(2018)Fan, Lewis, and Dauphin}]{fan2018hierarchical}
Angela Fan, Mike Lewis, and Yann Dauphin. 2018.
\newblock \href {https://aclanthology.org/P18-1082} {Hierarchical neural story
  generation}.
\newblock In \emph{Proceedings of ACL}, pages 889--898.

\bibitem[{Fleiss(1971)}]{Fleiss1971MeasuringNS}
Joseph~L Fleiss. 1971.
\newblock \href {https://psycnet.apa.org/record/1972-05083-001} {Measuring
  nominal scale agreement among many raters}.
\newblock \emph{Psychological Bulletin}, 76(5):378–382.

\bibitem[{Fu et~al.(2018)Fu, Tan, Peng, Zhao, and Yan}]{fu2018style}
Zhenxin Fu, Xiaoye Tan, Nanyun Peng, Dongyan Zhao, and Rui Yan. 2018.
\newblock \href {https://ojs.aaai.org/index.php/AAAI/article/view/11330} {Style
  transfer in text: Exploration and evaluation}.
\newblock In \emph{Proceedings of AAAI}, pages 663--670.

\bibitem[{Ghazvininejad et~al.(2019)Ghazvininejad, Levy, Liu, and
  Zettlemoyer}]{Ghazvininejad2019MaskPredictPD}
Marjan Ghazvininejad, Omer Levy, Yinhan Liu, and Luke Zettlemoyer. 2019.
\newblock \href {https://aclanthology.org/D19-1633/} {Mask-predict: Parallel
  decoding of conditional masked language models}.
\newblock In \emph{Proceedings of EMNLP}, pages 6112--6121.

\bibitem[{Goodfellow et~al.(2014)Goodfellow, Pouget-Abadie, Mirza, Xu,
  Warde-Farley, Ozair, Courville, and Bengio}]{goodfellowgenerative}
Ian~J Goodfellow, Jean Pouget-Abadie, Mehdi Mirza, Bing Xu, David Warde-Farley,
  Sherjil Ozair, Aaron Courville, and Yoshua Bengio. 2014.
\newblock \href
  {https://proceedings.neurips.cc/paper/2014/hash/5ca3e9b122f61f8f06494c97b1afccf3-Abstract.html}
  {Generative adversarial nets}.
\newblock In \emph{NIPS}, pages 2672--2680.

\bibitem[{Gu et~al.(2018)Gu, Bradbury, Xiong, Li, and
  Socher}]{Gu2018NonAutoregressiveNM}
Jiatao Gu, James Bradbury, Caiming Xiong, Victor~O.K. Li, and Richard Socher.
  2018.
\newblock \href {https://openreview.net/forum?id=B1l8BtlCb} {Non-autoregressive
  neural machine translation}.
\newblock In \emph{ICLR}.

\bibitem[{Gu et~al.(2019)Gu, Wang, and Zhao}]{gu2019levenshtein}
Jiatao Gu, Changhan Wang, and Jake Zhao. 2019.
\newblock \href
  {https://proceedings.neurips.cc/paper/2019/hash/675f9820626f5bc0afb47b57890b466e-Abstract.html}
  {Levenshtein transformer}.
\newblock In \emph{NIPS}, pages 11181--11191.

\bibitem[{He and Li(2021)}]{he2021xlentmcmc}
Xingwei He and Victor~O.K. Li. 2021.
\newblock \href {https://ojs.aaai.org/index.php/AAAI/article/view/17536} {Show
  me how to revise: Improving lexically constrained sentence generation with
  {XLNet}}.
\newblock In \emph{Proceedings of AAAI}, pages 12989--12997.

\bibitem[{Hokamp and Liu(2017)}]{Hokamp2017LexicallyCD}
Chris Hokamp and Qun Liu. 2017.
\newblock \href {https://aclanthology.org/P17-1141} {Lexically constrained
  decoding for sequence generation using grid beam search}.
\newblock In \emph{Proceedings of ACL}, pages 1535--1546.

\bibitem[{Holtzman et~al.(2020)Holtzman, Buys, Du, Forbes, and
  Choi}]{Holtzman2020TheCC}
Ari Holtzman, Jan Buys, Li~Du, Maxwell Forbes, and Yejin Choi. 2020.
\newblock \href {https://openreview.net/forum?id=rygGQyrFvH} {The curious case
  of neural text degeneration}.
\newblock In \emph{ICLR}.

\bibitem[{Holtzman et~al.(2018)Holtzman, Buys, Forbes, Bosselut, Golub, and
  Choi}]{Holtzman2018LearningTW}
Ari Holtzman, Jan Buys, Maxwell Forbes, Antoine Bosselut, David Golub, and
  Yejin Choi. 2018.
\newblock \href {https://aclanthology.org/P18-1152/} {Learning to write with
  cooperative discriminators}.
\newblock In \emph{Proceedings of ACL}, pages 1638--1649.

\bibitem[{Huang et~al.(2020)Huang, Zhang, Elachqar, and
  Cheng}]{huang-etal-2020-inset}
Yichen Huang, Yizhe Zhang, Oussama Elachqar, and Yu~Cheng. 2020.
\newblock \href {https://aclanthology.org/2020.acl-main.226/} {{INSET}:
  Sentence infilling with {IN}ter-{SE}ntential transformer}.
\newblock In \emph{Proceedings of ACL}, pages 2502--2515.

\bibitem[{Keskar et~al.(2019)Keskar, McCann, Varshney, Xiong, and
  Socher}]{Keskar2019CTRLAC}
Nitish~Shirish Keskar, Bryan McCann, Lav~R Varshney, Caiming Xiong, and Richard
  Socher. 2019.
\newblock \href {https://arxiv.org/abs/1909.05858} {{CTRL}: A conditional
  transformer language model for controllable generation}.
\newblock \emph{arXiv preprint arXiv:1909.05858}.

\bibitem[{Landis and Koch(1977)}]{Landis1977TheMO}
J~Richard Landis and Gary~G Koch. 1977.
\newblock \href
  {https://www.jstor.org/stable/2529310?seq=1#metadata_info_tab_contents} {The
  measurement of observer agreement for categorical data}.
\newblock \emph{Biometrics}, 33(1):159--174.

\bibitem[{Lee et~al.(2018)Lee, Mansimov, and Cho}]{Lee2018DeterministicNN}
Jason Lee, Elman Mansimov, and Kyunghyun Cho. 2018.
\newblock \href {https://aclanthology.org/D18-1149} {Deterministic
  non-autoregressive neural sequence modeling by iterative refinement}.
\newblock In \emph{Proceedings of EMNLP}, pages 1173--1182.

\bibitem[{Lewis et~al.(2020)Lewis, Liu, Goyal, Ghazvininejad, Mohamed, Levy,
  Stoyanov, and Zettlemoyer}]{Lewis2020BARTDS}
Mike Lewis, Yinhan Liu, Naman Goyal, Marjan Ghazvininejad, Abdelrahman Mohamed,
  Omer Levy, Ves Stoyanov, and Luke Zettlemoyer. 2020.
\newblock \href {https://aclanthology.org/2020.acl-main.703/} {{BART}:
  Denoising sequence-to-sequence pre-training for natural language generation,
  translation, and comprehension}.
\newblock In \emph{Proceedings of ACL}, pages 7871--7880.

\bibitem[{Li et~al.(2016)Li, Galley, Brockett, Gao, and Dolan}]{Li2016ADO}
Jiwei Li, Michel Galley, Chris Brockett, Jianfeng Gao, and Bill Dolan. 2016.
\newblock \href {https://aclanthology.org/N16-1014} {A diversity-promoting
  objective function for neural conversation models}.
\newblock In \emph{Proceedings of NAACL-HLT}, pages 110--119.

\bibitem[{Li et~al.(2018)Li, Jia, He, and Liang}]{li2018delete}
Juncen Li, Robin Jia, He~He, and Percy Liang. 2018.
\newblock \href {https://aclanthology.org/N18-1169/} {Delete, retrieve,
  generate: A simple approach to sentiment and style transfer}.
\newblock In \emph{Proceedings of NAACL-HLT}, pages 1865--1874.

\bibitem[{Lin et~al.(2020)Lin, Zhou, Shen, Zhou, Bhagavatula, Choi, and
  Ren}]{lin-etal-2020-commongen}
Bill~Yuchen Lin, Wangchunshu Zhou, Ming Shen, Pei Zhou, Chandra Bhagavatula,
  Yejin Choi, and Xiang Ren. 2020.
\newblock \href {https://aclanthology.org/2020.findings-emnlp.165}
  {{C}ommon{G}en: A constrained text generation challenge for generative
  commonsense reasoning}.
\newblock In \emph{Findings of EMNLP}, pages 1823--1840.

\bibitem[{Liu et~al.(2019{\natexlab{a}})Liu, Fu, Qu, and Lv}]{Liu2019BFGANBA}
Dayiheng Liu, Jie Fu, Qian Qu, and Jiancheng Lv. 2019{\natexlab{a}}.
\newblock \href {https://ieeexplore.ieee.org/abstract/document/8846084}
  {{BFGAN}: Backward and forward generative adversarial networks for lexically
  constrained sentence generation}.
\newblock \emph{IEEE/ACM Transactions on Audio, Speech, and Language
  Processing}, 27(12):2350--2361.

\bibitem[{Liu et~al.(2019{\natexlab{b}})Liu, Ott, Goyal, Du, Joshi, Chen, Levy,
  Lewis, Zettlemoyer, and Stoyanov}]{Liu2019RoBERTaAR}
Yinhan Liu, Myle Ott, Naman Goyal, Jingfei Du, Mandar Joshi, Danqi Chen, Omer
  Levy, Mike Lewis, Luke Zettlemoyer, and Veselin Stoyanov. 2019{\natexlab{b}}.
\newblock \href {https://arxiv.org/abs/1907.11692} {{RoBERTa}: A robustly
  optimized {BERT} pretraining approach}.
\newblock \emph{arXiv preprint arXiv:1907.11692}.

\bibitem[{Loshchilov and Hutter(2019)}]{Loshchilov2019DecoupledWD}
Ilya Loshchilov and Frank Hutter. 2019.
\newblock \href {https://openreview.net/forum?id=Bkg6RiCqY7} {Decoupled weight
  decay regularization}.
\newblock In \emph{ICLR}.

\bibitem[{Miao et~al.(2019)Miao, Zhou, Mou, Yan, and Li}]{miao2019cgmh}
Ning Miao, Hao Zhou, Lili Mou, Rui Yan, and Lei Li. 2019.
\newblock \href {https://ojs.aaai.org/index.php/AAAI/article/view/4659}
  {{CGMH}: Constrained sentence generation by metropolis-hastings sampling}.
\newblock In \emph{Proceedings of AAAI}, pages 6834--6842.

\bibitem[{Mou et~al.(2016)Mou, Song, Yan, Li, Zhang, and Jin}]{mou2016}
Lili Mou, Yiping Song, Rui Yan, Ge~Li, Lu~Zhang, and Zhi Jin. 2016.
\newblock \href {https://aclanthology.org/C16-1316} {Sequence to backward and
  forward sequences: A content-introducing approach to generative short-text
  conversation}.
\newblock In \emph{Proceedings of COLING}, pages 3349--3358.

\bibitem[{Mou et~al.(2015)Mou, Yan, Li, Zhang, and Jin}]{mou2015backward}
Lili Mou, Rui Yan, Ge~Li, Lu~Zhang, and Zhi Jin. 2015.
\newblock \href {https://arxiv.org/abs/1512.06612} {Backward and forward
  language modeling for constrained sentence generation}.
\newblock \emph{arXiv preprint arXiv:1512.06612}.

\bibitem[{Papineni et~al.(2002)Papineni, Roukos, Ward, and
  Zhu}]{Papineni2002BleuAM}
Kishore Papineni, Salim Roukos, Todd Ward, and Wei-Jing Zhu. 2002.
\newblock \href {https://aclanthology.org/P02-1040} {{BLEU}: A method for
  automatic evaluation of machine translation}.
\newblock In \emph{Proceedings of ACL}, pages 311--318.

\bibitem[{Post and Vilar(2018)}]{post-vilar-2018-fast}
Matt Post and David Vilar. 2018.
\newblock \href {https://aclanthology.org/N18-1119} {Fast lexically constrained
  decoding with dynamic beam allocation for neural machine translation}.
\newblock In \emph{Proceedings of NAACL-HLT}, pages 1314--1324.

\bibitem[{Radford(2018)}]{Radford2018ImprovingLU}
Alec Radford. 2018.
\newblock \href
  {https://www.cs.ubc.ca/~amuham01/LING530/papers/radford2018improving.pdf}
  {Improving language understanding by generative pre-training}.
\newblock \emph{OpenAI Technical Report}.

\bibitem[{Radford et~al.(2019)Radford, Wu, Child, Luan, Amodei, and
  Sutskever}]{Radford2019LanguageMA}
Alec Radford, Jeffrey Wu, Rewon Child, David Luan, Dario Amodei, and Ilya
  Sutskever. 2019.
\newblock \href {http://www.persagen.com/files/misc/radford2019language.pdf}
  {Language models are unsupervised multitask learners}.
\newblock \emph{OpenAI Technical Report}.

\bibitem[{Sha(2020)}]{sha-2020-gradient}
Lei Sha. 2020.
\newblock \href {https://aclanthology.org/2020.emnlp-main.701} {Gradient-guided
  unsupervised lexically constrained text generation}.
\newblock In \emph{Proceedings of EMNLP}, pages 8692--8703.

\bibitem[{Shen et~al.(2017)Shen, Lei, Barzilay, and Jaakkola}]{shen2017style}
Tianxiao Shen, Tao Lei, Regina Barzilay, and Tommi Jaakkola. 2017.
\newblock \href
  {https://proceedings.neurips.cc/paper/2017/hash/2d2c8394e31101a261abf1784302bf75-Abstract.html}
  {Style transfer from non-parallel text by cross-alignment}.
\newblock In \emph{NIPS}, pages 6830--6841.

\bibitem[{Stern et~al.(2019)Stern, Chan, Kiros, and
  Uszkoreit}]{Stern2019InsertionTF}
Mitchell Stern, William Chan, J.~Kiros, and Jakob Uszkoreit. 2019.
\newblock \href {http://proceedings.mlr.press/v97/stern19a.html} {Insertion
  transformer: Flexible sequence generation via insertion operations}.
\newblock In \emph{ICML}, pages 5976--5985.

\bibitem[{Su et~al.(2018)Su, Xu, Qiu, and Huang}]{su2018incorporating}
Jinyue Su, Jiacheng Xu, Xipeng Qiu, and Xuanjing Huang. 2018.
\newblock \href {https://ojs.aaai.org/index.php/AAAI/article/view/11990}
  {Incorporating discriminator in sentence generation: a gibbs sampling
  method}.
\newblock In \emph{Proceedings of AAAI}, pages 5496--5503.

\bibitem[{Susanto et~al.(2020)Susanto, Chollampatt, and
  Tan}]{susanto-etal-2020-lexically}
Raymond~Hendy Susanto, Shamil Chollampatt, and Liling Tan. 2020.
\newblock \href {https://aclanthology.org/2020.acl-main.325} {Lexically
  constrained neural machine translation with {L}evenshtein transformer}.
\newblock In \emph{Proceedings of ACL}, pages 3536--3543.

\bibitem[{Tan et~al.(2021)Tan, Yang, Al-Shedivat, Xing, and
  Hu}]{tan2020progressive}
Bowen Tan, Zichao Yang, Maruan Al-Shedivat, Eric Xing, and Zhiting Hu. 2021.
\newblock \href {https://aclanthology.org/2021.naacl-main.341} {Progressive
  generation of long text with pretrained language models}.
\newblock In \emph{Proceedings of NAACL-HLT}, pages 4313--4324.

\bibitem[{Wolf et~al.(2019)Wolf, Debut, Sanh, Chaumond, Delangue, Moi, Cistac,
  Rault, Louf, Funtowicz, and Brew}]{Wolf2019HuggingFacesTS}
Thomas Wolf, Lysandre Debut, Victor Sanh, Julien Chaumond, Clement Delangue,
  Anthony Moi, Pierric Cistac, Tim Rault, R'emi Louf, Morgan Funtowicz, and
  Jamie Brew. 2019.
\newblock \href {https://arxiv.org/abs/1910.03771} {Huggingface's transformers:
  State-of-the-art natural language processing}.
\newblock \emph{arXiv preprint arXiv:1910.03771}.

\bibitem[{Xu et~al.(2018)Xu, Sun, Zeng, Zhang, Ren, Wang, and
  Li}]{xu2018unpaired}
Jingjing Xu, Xu~Sun, Qi~Zeng, Xiaodong Zhang, Xuancheng Ren, Houfeng Wang, and
  Wenjie Li. 2018.
\newblock \href {https://aclanthology.org/P18-1090} {Unpaired
  sentiment-to-sentiment translation: A cycled reinforcement learning
  approach}.
\newblock In \emph{Proceedings of ACL}, pages 979--988.

\bibitem[{Yang et~al.(2019)Yang, Dai, Yang, Carbonell, Salakhutdinov, and
  Le}]{Yang2019XLNetGA}
Zhilin Yang, Zihang Dai, Yiming Yang, Jaime~G. Carbonell, Ruslan Salakhutdinov,
  and Quoc~V. Le. 2019.
\newblock \href
  {https://proceedings.neurips.cc/paper/2019/hash/dc6a7e655d7e5840e66733e9ee67cc69-Abstract.html}
  {{XLNet}: Generalized autoregressive pretraining for language understanding}.
\newblock In \emph{NIPS}, pages 5753--5763.

\bibitem[{Zhang et~al.(2020{\natexlab{a}})Zhang, Jiang, Li, and
  Xue}]{zhang2020languagegeneration}
Maosen Zhang, Nan Jiang, Lei Li, and Yexiang Xue. 2020{\natexlab{a}}.
\newblock \href {https://aclanthology.org/2020.findings-emnlp.115} {Language
  generation via combinatorial constraint satisfaction: A tree search enhanced
  {M}onte-{C}arlo approach}.
\newblock In \emph{Findings of EMNLP}, pages 1286--1298.

\bibitem[{Zhang et~al.(2020{\natexlab{b}})Zhang, Wang, Li, Gan, Brockett, and
  Dolan}]{zhang2020pointer}
Yizhe Zhang, Guoyin Wang, Chunyuan Li, Zhe Gan, Chris Brockett, and Bill Dolan.
  2020{\natexlab{b}}.
\newblock \href {https://aclanthology.org/2020.emnlp-main.698} {{POINTER}:
  Constrained progressive text generation via insertion-based generative
  pre-training}.
\newblock In \emph{Proceedings of EMNLP}, pages 8649--8670.

\bibitem[{Zhu et~al.(2018)Zhu, Lu, Zheng, Guo, Zhang, Wang, and
  Yu}]{Zhu2018TexygenAB}
Yaoming Zhu, Sidi Lu, Lei Zheng, Jiaxian Guo, Weinan Zhang, Jun Wang, and Yong
  Yu. 2018.
\newblock \href {https://dl.acm.org/doi/abs/10.1145/3209978.3210080} {Texygen:
  A benchmarking platform for text generation models}.
\newblock In \emph{SIGIR}, pages 1097--1100.

\end{thebibliography}

\clearpage
\appendix{\textbf{Appendix}}
\section{Performance of Language Models}\label{appendix_a}
The forward GPT-2, backward GPT-2, separate forward GPT-2 and 
separate backward GPT-2 are initialized with the pre-trained GPT-2 small model. 
These models are fine-tuned on the training sets of One-Billion-Word or Yelp. 
We choose the checkpoint with the lowest NLL loss on the validation set.
They are used for baselines, 
including sep-B/F, asyn-B/F, GBS and CGMH. 
NLL results of the fine-tuned language models on the validation sets are shown in Table \ref{tab:appendix_nll}.

\section{Performance of Classifiers}\label{appendix_b}
We create synthetic datasets for CBART with One-Billion-Word and Yelp. 
We select $1M$ and $0.1M$ sentences from each dataset as the 
training and validation sets. For each sentence, we create 10 synthetic data instances. 
Therefore, for each dataset, the size of the synthetic training and validation sets are $10M$ and $1M$. 
We fine-tune CBART on the synthetic training set for two epochs with a learning rate of $1e-5$ 
and select the best checkpoint with the lowest loss on the synthetic validation set. 
We show the performance of classifiers of CBART-base (the base CBART model, which is initialized with BART-base.) and CBART-large 
(the large CBART model, which is initialized with BART-large.) 
in Table \ref{tab:appendix_classifier1} and Table \ref{tab:appendix_classifier2}, respectively.  

\section{Generating Text with Lexical Constraints}\label{appendix_d}
We show some text generated by 
baselines and our proposed model with lexical constraints extracted from 
One-Billion-Word and Yelp test sets in Table \ref{tab:case2} and Table \ref{tab:case3}, respectively.

\begin{table}[h] 
  \footnotesize
    \centering
      \begin{tabular}{
      m{0.21\textwidth}<{\centering}|
       m{0.15\textwidth}<{\centering}
       m{0.04\textwidth}<{\centering}
       }
      \toprule
       Models & One-Billion-Word& Yelp\\
       \midrule
       Forward GPT-2  & 3.463 & 2.885 \\
       Backward GPT-2 & 4.130 & 3.171 \\
       Separate forward GPT-2  & 3.540 & 3.055 \\
       Separate backward GPT-2 & 4.377 & 3.260 \\
      \bottomrule
    \end{tabular}
    \caption{NLL of different language models on the validation sets. 
    }\label{tab:appendix_nll} 
  \end{table}

\begin{table}[t] 
\footnotesize
  \centering
    \begin{tabular}{
      m{0.135\textwidth}<{\centering}|
    m{0.12\textwidth}<{\centering}|
     m{0.03\textwidth}<{\centering}
     m{0.03\textwidth}<{\centering}
     m{0.03\textwidth}<{\centering}
     }
    \toprule
     Datasets&Labels & P &R & F1 \\
     \midrule
     \multirow{4}{*}{One-Billion-Word}&Copy  &0.970	&0.985	&0.978 \\
     &Replacement  &0.981	&0.975	&0.978\\
     &Insertion  &0.928	&0.862	&0.894\\
     &Macro-average &0.960	&0.941	&0.950 \\  
     \hline
     \multirow{4}{*}{Yelp}&Copy &0.974	&0.985	&0.979  \\
     &Replacement &0.993	&0.991	&0.992 \\
     &Insertion &0.928	&0.874	&0.900 \\
     &Macro-average &0.965	&0.950	&0.957 \\   
    \bottomrule
  \end{tabular}
  \caption{Results of the classifier of CBART-base on the synthetic validation sets of One-Billion-Word and Yelp. 
  ``P'' and ``R'' denote precision and recall. 
  }\label{tab:appendix_classifier1}
\end{table}

\begin{table}[t]
\footnotesize
  \centering
    \begin{tabular}{
      m{0.135\textwidth}<{\centering}|
    m{0.12\textwidth}<{\centering}|
     m{0.03\textwidth}<{\centering}
     m{0.03\textwidth}<{\centering}
     m{0.03\textwidth}<{\centering}
     }
    \toprule
     Datasets&Labels & P &R & F1 \\
     \midrule
     \multirow{4}{*}{One-Billion-Word}&Copy &0.974 & 0.986 & 0.980  \\
     &Replacement &0.985	&0.984	&0.984 \\
     &Insertion &0.933	&0.880	&0.906 \\
     &Macro-average &0.964	&0.950	&0.957 \\  
     \hline
     \multirow{4}{*}{{Yelp}}&Copy  &0.978	&0.985	&0.981 \\
     &Replacement  &0.995	&0.994	&0.994\\
     &Insertion  &0.925	&0.894	&0.910\\
     &Macro-average &0.966	&0.958	&0.962 \\    
    \bottomrule
  \end{tabular}
  \caption{Results of the classifier of CBART-large on the synthetic validation sets of One-Billion-Word and Yelp. 
  ``P'' and ``R'' denote precision and recall. 
  }\label{tab:appendix_classifier2}
\end{table}

\begin{table*}[t]
  \footnotesize
  \centering
  \begin{tabular}{
    m{0.01\textwidth}<{}
    m{0.11\textwidth}<{\raggedright}|
    m{0.81\textwidth}<{}
    }
    \toprule
    &\textbf{Constraints} & \textbf{hearing}, \textbf{system}, \textbf{need}\\
    \midrule
    &Human& We are already \textbf{hearing} arguments for focusing everything on the economy damaged by failure in the banking \textbf{system} , dropping the \textbf{need} to fix the climate system .	 \\
    \midrule
    \multirow{5}{*}{\rotatebox{90}{\textbf{Baselines}}}&GBS & `` We \textbf{need} to have a \textbf{system} of \textbf{hearing} protection , " he said .\\
    \cline{2-3} 
    &CGMH& A public \textbf{hearing} is just one more \textbf{system} that we \textbf{need} .		\\
    \cline{2-3} 
    &X-MCMC-C& The new public \textbf{hearing} \textbf{system} will \textbf{need} funding for about six months after the court ruling .  \\
    \cline{2-3} 
    &POINTER-2& and he said at the senate \textbf{hearing} that it ? s changing in the entire current \textbf{system} , it ? s a \textbf{need} for a reform now . . . \\
    \midrule
    \multirow{5}{*}{\rotatebox{90}{\textbf{CBART}}}&\textbf{greedy}& The new \textbf{hearing} \textbf{system} is less expensive , and there was no \textbf{need} for a specialist .\\
    \cline{2-3} 
    &\textbf{$k$=5, $c$=1}&`` The \textbf{hearing} \textbf{system} is something we \textbf{need} to improve .\\
    \cline{2-3} 
    &\textbf{$k$=50, $c$=1}&She has a \textbf{hearing} \textbf{system} and is in \textbf{need} of glasses .	\\
    \cline{2-3}
    &\textbf{$p$=0.5, $c$=1}&The \textbf{hearing} \textbf{system} is closed , but you will not \textbf{need} to pay for it .	\\
    \cline{2-3}
    &\textbf{$p$=0.9, $c$=1}&Is there any existing \textbf{hearing} \textbf{system} that would \textbf{need} to be adapted ?	\\
    \midrule
    \midrule
    
    &\textbf{Constraints} & \textbf{admitted}, \textbf{health}, \textbf{heavy}, \textbf{cold}\\
    \midrule
    &Human& Philip , 86 , was \textbf{admitted} to the King Edward VII Hospital in central London on Thursday after his \textbf{health} deteriorated having caught a \textbf{heavy} \textbf{cold} . \\
    \midrule
    \multirow{6}{*}{\rotatebox{90}{\textbf{Baselines}}}&GBS & He \textbf{admitted} that he had been `` \textbf{cold} and \textbf{heavy} " in the \textbf{health} department	\\
    \cline{2-3}
    &CGMH& He \textbf{admitted} that he had faced mental \textbf{health} problems and a \textbf{heavy} \textbf{cold} .	 \\
    \cline{2-3}
    &X-MCMC-C& The Labour MP was \textbf{admitted} to a mental \textbf{health} hospital after suffering a \textbf{heavy} \textbf{cold} for most of the week .\\
    \cline{2-3}
    &POINTER-2& but when she was \textbf{admitted} to a local mental \textbf{health} unit because what she did was so \textbf{heavy} a burden on mea , and one or more afraid of the other colds . . . ? \\
    \midrule
    \multirow{7}{*}{\rotatebox{90}{\textbf{CBART}}}&\textbf{greedy}& The singer \textbf{admitted} to having mental \textbf{health} problems and suffering from a \textbf{heavy} \textbf{cold} .	 \\
    \cline{2-3}
    &\textbf{$k$=5, $c$=1}&The court \textbf{admitted} the pair have two previous \textbf{health} problems , including \textbf{heavy} doses of \textbf{cold} and flu medication .	    \\
    \cline{2-3}
    &\textbf{$k$=50, $c$=1}&Health ministers \textbf{admitted} that the \textbf{health} service was weak because it had suffered a \textbf{heavy} dose of \textbf{cold} and flu .	    \\
    \cline{2-3}
    &\textbf{$p$=0.5, $c$=1}&She \textbf{admitted} to having \textbf{health} problems , including \textbf{heavy} \textbf{cold} and flu .			\\
    \cline{2-3}
    &\textbf{$p$=0.9, $c$=1}&He had been \textbf{admitted} to the hospital mental \textbf{health} unit , suffering from a \textbf{heavy} head \textbf{cold} and fever .	    \\
    
    \midrule
    \midrule
    
    &\textbf{Constraints} & \textbf{way}, \textbf{back}, \textbf{missing}, \textbf{weeks}, \textbf{due}\\
    \midrule
    &Human& John Lackey is all the \textbf{way} \textbf{back} after \textbf{missing} the first six \textbf{weeks} \textbf{due} to injury , and pitching like an ace again .	  \\
    \midrule
    \multirow{5}{*}{\rotatebox{90}{\textbf{Baselines}}}&GBS & `` The \textbf{way} \textbf{back} is \textbf{due} to the \textbf{missing} \textbf{weeks} , " he said .	    \\
    \cline{2-3}
    &CGMH& The only \textbf{way} \textbf{back} is after \textbf{missing} four \textbf{weeks} \textbf{due} to injury .	    \\
    \cline{2-3}
    &X-MCMC-C& He is on his \textbf{way} \textbf{back} home after \textbf{missing} two \textbf{weeks} \textbf{due} to an Achilles tendon injury .\\
    \cline{2-3}
    &POINTER-2& he has found his \textbf{way} into england despite his \textbf{back} injury , \textbf{missing} three of the last two \textbf{weeks} with a calf injury and another two \textbf{due} to a calf injury . \\
    \midrule
    \multirow{5}{*}{\rotatebox{90}{\textbf{CBART}}}&\textbf{greedy}& He is on his \textbf{way} \textbf{back} to the squad after \textbf{missing} two \textbf{weeks} \textbf{due} to a stomach bug . \\
    \cline{2-3}
    &\textbf{$k$=5, $c$=1}&She is making her \textbf{way} \textbf{back} into action after \textbf{missing} three \textbf{weeks} \textbf{due} to illness .	    \\
    \cline{2-3}
    &\textbf{$k$=50, $c$=1}&Took his \textbf{way} \textbf{back} from a groin injury after \textbf{missing} six \textbf{weeks} \textbf{due} to knee injuries .	    \\
    \cline{2-3}
    &\textbf{$p$=0.5, $c$=1}&But he was on his \textbf{way} \textbf{back} after \textbf{missing} two \textbf{weeks} \textbf{due} to a visa issue .	    \\
    \cline{2-3}
    &\textbf{$p$=0.9, $c$=1}&He is on the \textbf{way} \textbf{back} after \textbf{missing} two \textbf{weeks} \textbf{due} to a fractured collarbone and bruised ribs .	\\
    \midrule
    \midrule
    
    &\textbf{Constraints} & \textbf{likely}, \textbf{certain}, \textbf{respond}, \textbf{others}, \textbf{new}, \textbf{environment}\\
    \midrule
    &Human& That said , it is \textbf{likely} that \textbf{certain} forms of religion would \textbf{respond} better than \textbf{others} to the \textbf{new} \textbf{environment} .	    \\
    \midrule
    \multirow{6}{*}{\rotatebox{90}{\textbf{Baselines}}}&GBS & `` The \textbf{new} \textbf{environment} is \textbf{likely} to \textbf{respond} to \textbf{certain} \textbf{others} , " he said	    \\
    \cline{2-3}
    &CGMH& Not \textbf{likely} , but \textbf{certain} areas will \textbf{respond} better than \textbf{others} to the \textbf{new} \textbf{environment} .	    \\
    \cline{2-3}
    &X-MCMC-C& And they are also \textbf{likely} to be \textbf{certain} to \textbf{respond} to \textbf{others} who want to create a \textbf{new} \textbf{environment} to replace the old . \\
    \cline{2-3}
    &POINTER-2& they are more \textbf{likely} to be unable to do a \textbf{certain} things , about how they \textbf{respond} better than \textbf{others} in this area , are new \textbf{new} or why the current \textbf{environment} is different .\\
    \midrule
    \multirow{7}{*}{\rotatebox{90}{\textbf{CBART}}}&\textbf{greedy}& The first group is most \textbf{likely} to be \textbf{certain} to \textbf{respond} better than \textbf{others} in a \textbf{new} \textbf{environment} .	    \\
    \cline{2-3}
    &\textbf{$k$=5, $c$=1}&The company is \textbf{likely} to depend on \textbf{certain} sectors and \textbf{respond} to \textbf{others} in the \textbf{new} \textbf{environment} .	    \\
    \cline{2-3}
    &\textbf{$k$=50, $c$=1}&We have no more \textbf{likely} chance of survival , while \textbf{certain} groups \textbf{respond} differently and the \textbf{others} adapt to a \textbf{new} \textbf{environment} .	    \\
    \cline{2-3}
    &\textbf{$p$=0.5, $c$=1}&How \textbf{likely} to be that you learn about \textbf{certain} issues and \textbf{respond} to \textbf{others} in a \textbf{new} \textbf{environment} .	    \\
    \cline{2-3}
    &\textbf{$p$=0.9, $c$=1}&I suspect it is \textbf{likely} to find \textbf{certain} chemicals would \textbf{respond} better than \textbf{others} , which could cope in the \textbf{new} \textbf{environment} .	    \\
    \bottomrule
  \end{tabular}
  \caption{Generated text with constraints from One-Billion-Word test sets. ``Human'' refers to the human reference.}\label{tab:case2}
\end{table*}

\begin{table*}[t]
  \footnotesize
  \centering
  \begin{tabular}{
    m{0.01\textwidth}<{}
    m{0.11\textwidth}<{\raggedright}|
    m{0.81\textwidth}<{}
  }
    \toprule
    &\textbf{Constraints} & \textbf{past}, \textbf{decided}, \textbf{try}\\
    \midrule
    &Human& i have driven \textbf{past} this place a few times , and finally one morning i \textbf{decided} to give it a \textbf{try} .	 \\
    \midrule
    \multirow{4}{*}{\rotatebox{90}{\textbf{Baselines}}}&GBS & i \textbf{decided} to give this place a \textbf{try} based on the \textbf{past} reviews .	\\
    \cline{2-3}
    &CGMH& driving \textbf{past} this place and \textbf{decided} to \textbf{try} it out .	\\
    \cline{2-3}
    &X-MCMC-C& we stayed here this \textbf{past} weekend and \textbf{decided} to \textbf{try} it out .\\
    \cline{2-3}
    &POINTER-2 & i have been walking \textbf{past} by here a few times , and finally \textbf{decided} to give it a \textbf{try} out . \\
    \midrule
    \multirow{6}{*}{\rotatebox{90}{\textbf{CBART}}}&\textbf{greedy}& i drive \textbf{past} this place everyday and finally \textbf{decided} to \textbf{try} it out .	\\
    \cline{2-3}
    &\textbf{$k$=5, $c$=1}&i walked \textbf{past} this restaurant and \textbf{decided} to give it a \textbf{try} one evening .	\\
    \cline{2-3}
    &\textbf{$k$=50, $c$=1}&we have been driving \textbf{past} this buffet but it was a while so we \textbf{decided} to \textbf{try} this restaurant since they were in a good location .	\\
    \cline{2-3}
    &\textbf{$p$=0.5, $c$=1}&i have driven \textbf{past} this place many times and finally \textbf{decided} to \textbf{try} it out .	\\
    \cline{2-3}
    &\textbf{$p$=0.95, $c$=1}&my husband has always drove \textbf{past} this place and we \textbf{decided} to \textbf{try} it .	\\

    \midrule
    \midrule
    
    &\textbf{Constraints} & \textbf{dinner}, \textbf{called}, \textbf{arrived}, \textbf{early}\\
    \midrule
    &Human& christmas eve \textbf{dinner} at a so \textbf{called} steakhouse . \textbf{arrived} \textbf{early} for our 630 reservation and we had to wait 20 minutes for a table .		 \\
    \midrule
    \multirow{5}{*}{\rotatebox{90}{\textbf{Baselines}}}&GBS & we \textbf{arrived} \textbf{early} for \textbf{dinner} and \textbf{called} to make an appointment .	\\
    \cline{2-3}
    &CGMH& when our first wedding anniversary \textbf{dinner} was \textbf{called} , the driver \textbf{arrived} 10 minutes \textbf{early} .	\\
    \cline{2-3}
    &X-MCMC-C& we just had \textbf{dinner} here tonight and we \textbf{called} ahead to place our order and \textbf{arrived} \textbf{early}  .\\
    \cline{2-3}
    &POINTER-2& went here to have \textbf{dinner} out here tonight . we had \textbf{called} in a few other places were ahead of me , and \textbf{arrived} just a few minutes \textbf{early} , and were promptly seated .\\
    \midrule
    \multirow{5}{*}{\rotatebox{90}{\textbf{CBART}}}&\textbf{greedy}& we had a wonderful \textbf{dinner} here last night . we \textbf{called} ahead and \textbf{arrived} 15 minutes \textbf{early} .	\\
    \cline{2-3}
    &\textbf{$k$=5, $c$=1}&i had \textbf{dinner} there last night . i \textbf{called} ahead for a reservation and \textbf{arrived} 10 minutes \textbf{early} .	\\
    \cline{2-3}
    &\textbf{$k$=50, $c$=1}&we came here with a \textbf{dinner} for 10 , \textbf{called} in and \textbf{arrived} 20 minutes \textbf{early} .	\\
    \cline{2-3}
    &\textbf{$p$=0.5, $c$=1}&had \textbf{dinner} reservations and \textbf{called} to confirm , \textbf{arrived} 15 minutes \textbf{early} .		\\
    \cline{2-3}
    &\textbf{$p$=0.95, $c$=1}&we had \textbf{dinner} here and enjoyed everything . i \textbf{called} in ahead so we \textbf{arrived} 15 minutes \textbf{early} .	\\
    
    \midrule
    \midrule
    
    &\textbf{Constraints} & \textbf{eat}, \textbf{crab}, \textbf{fresh}, \textbf{large}, \textbf{salty}\\
    \midrule
    &Human& got all u can \textbf{eat} \textbf{crab} legs and they were \textbf{fresh} , \textbf{large} , not \textbf{salty} , and not over cooked like some local restaurants .		 \\
    \midrule
    \multirow{5}{*}{\rotatebox{90}{\textbf{Baselines}}}&GBS & this is a great place to \textbf{eat} . \textbf{fresh} \textbf{crab} , \textbf{large} \textbf{salty} fish ,		\\
    \cline{2-3}
    &CGMH& do not \textbf{eat} the \textbf{crab} legs here ! \textbf{fresh} , \textbf{large} , and \textbf{salty} !	\\
    \cline{2-3}
    &X-MCMC-C&the best all you can \textbf{eat} buffet the \textbf{crab} legs are \textbf{fresh} and \textbf{large} and everything is not too \textbf{salty} or heavy . \\
    \cline{2-3}
    &POINTER-2& great place to \textbf{eat} at off the strip ! i had the snow \textbf{crab} wontons . they were \textbf{fresh} and delicious . they were in a very \textbf{large} portion and were not greasy or too \textbf{salty} at all .\\
    \midrule
    \multirow{7}{*}{\rotatebox{90}{\textbf{CBART}}}&\textbf{greedy}& i \textbf{eat} here often , the \textbf{crab} legs are \textbf{fresh} and \textbf{large} and not too \textbf{salty} .	\\
    \cline{2-3}
    &\textbf{$k$=5, $c$=1}&this is a place i \textbf{eat} at ! the \textbf{crab} cakes were \textbf{fresh} , \textbf{large} and not overly \textbf{salty} .		\\
    \cline{2-3}
    &\textbf{$k$=50, $c$=1}&this is one great place to \textbf{eat} at on vacation ! the \textbf{crab} legs were \textbf{fresh} and \textbf{large} slices of butterfish ; very \textbf{salty} , though .\\
    \cline{2-3}
    &\textbf{$p$=0.5, $c$=1}&all you can \textbf{eat} here is great . the \textbf{crab} legs were \textbf{fresh} and \textbf{large} , but a little too \textbf{salty} .	\\
    \cline{2-3}
    &\textbf{$p$=0.95, $c$=1}&we came just for the opportunity of hicklate late night bite too \textbf{eat} here . oysters \& \textbf{crab} legs are \textbf{fresh} , \textbf{large} portion and \textbf{salty} enough to finish !\\
    \midrule
    \midrule
    
    &\textbf{Constraints} & \textbf{way}, \textbf{home}, \textbf{decided}, \textbf{friendly}, \textbf{loved}, \textbf{atmosphere}\\
    \midrule
    &Human& this place was on the \textbf{way} back \textbf{home} for me and i \textbf{decided} to try it out . the workers were super \textbf{friendly} and i  \textbf{loved} the \textbf{atmosphere} .		 \\
    \midrule
    \multirow{6}{*}{\rotatebox{90}{\textbf{Baselines}}}&GBS & i  \textbf{loved} the \textbf{atmosphere} and \textbf{friendly} staff . i \textbf{decided} to go \textbf{home} \textbf{way} before		\\
    \cline{2-3}
    &CGMH& our \textbf{way} \textbf{home} was \textbf{decided} very fast ! very \textbf{friendly} service ,  \textbf{loved} the food and \textbf{atmosphere} !	\\
    \cline{2-3}
    &X-MCMC-C& it was on our \textbf{way} \textbf{home} we \textbf{decided} to stop in and they were very \textbf{friendly} and we \textbf{loved} the \textbf{atmosphere} and decor .\\
    \cline{2-3}
    &POINTER-2& we were out on our \textbf{way} back to \textbf{home} , and we are so very glad that we \textbf{decided} to stop here . very \textbf{friendly} local coffee shop ! i \textbf{loved} the decor and the \textbf{atmosphere} of this place ! .\\
    \midrule
    \multirow{7}{*}{\rotatebox{90}{\textbf{CBART}}}&\textbf{ greedy}& we were on our \textbf{way} \textbf{home} and \textbf{decided} to stop for a bite , \textbf{friendly} staff and  \textbf{loved} the \textbf{atmosphere} .	\\
    \cline{2-3}
    &\textbf{$k$=5, $c$=1}&on my \textbf{way} \textbf{home} and \textbf{decided} to try it out ! \textbf{friendly} staff ,  \textbf{loved} the \textbf{atmosphere} .		\\
    \cline{2-3}
    &\textbf{$k$=50, $c$=1}&we saw it on our \textbf{way} \textbf{home} and \textbf{decided} to stop in . very \textbf{friendly} staff and i  \textbf{loved} the \textbf{atmosphere} of this brewery !	\\
    \cline{2-3}
    &\textbf{$p$=0.5, $c$=1}&on our \textbf{way} \textbf{home} and \textbf{decided} to stop for dinner ! \textbf{friendly} staff ,  \textbf{loved} the \textbf{atmosphere} .\\
    \cline{2-3}
    &\textbf{$p$=0.95, $c$=1}&stumbled upon this place on my \textbf{way} \textbf{home} one night , \& \textbf{decided} to give it a try ! super \textbf{friendly} staff and  \textbf{loved} the \textbf{atmosphere} .	\\
    \bottomrule
  \end{tabular}
  \caption{Generated text with constraints from Yelp test sets. ``Human'' refers to the human reference.}\label{tab:case3}
\end{table*}

\end{document}